\documentclass[sigconf, screen, nonacm]{acmart}
\AtBeginDocument{%
  }

\setcopyright{acmlicensed}
\copyrightyear{2026}
\acmYear{2026}
\acmDOI{XXXXXXX.XXXXXXX}
\acmISBN{978-1-4503-XXXX-X/2018/06}

\usepackage{comment}
\usepackage{enumitem}
\usepackage{makecell} 
\usepackage{hyperref} 
\usepackage{colortbl}
\usepackage{multirow}
\usepackage{xcolor}
\usepackage{pifont}
\usepackage{graphicx}

\definecolor{yellowbg}{RGB}{255, 255, 204} 
\definecolor{orangebg}{RGB}{255, 229, 204} 
\definecolor{redbg}{RGB}{255, 204, 204}    
\definecolor{lightblue}{RGB}{224, 255, 255}

\newcommand{\cmark}{\textcolor{green!60!black}{\ding{51}}}
\newcommand{\xmark}{\textcolor{red!70!black}{\ding{55}}}
\begin{document}

\title{%
  \raisebox{-0.3\height}{\includegraphics[height=1.5em]{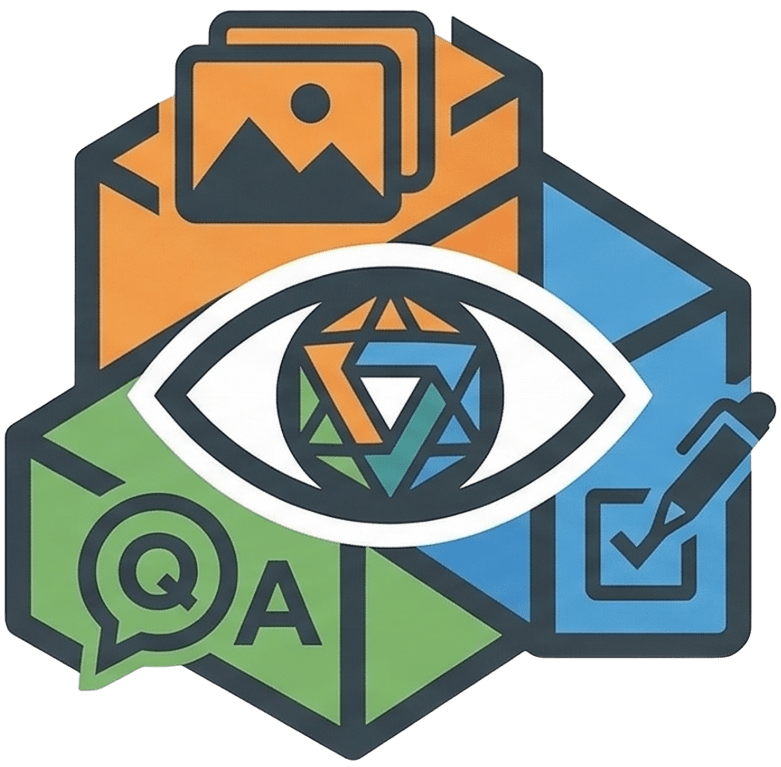}}%
  \,
  InsightVQA: High-Dimensional Emotion-Cognitive Visual Question Answering Benchmark%
}
\author{Shiyu Wang}
\authornote{Both authors contributed equally to this research.}
\author{Ziyu Liu}
\authornotemark[1]
\affiliation{%
  \institution{East China Normal University}
  \city{Shanghai}
  \country{China}
}
\email{51285903056@stu.ecnu.edu.cn}
\email{10225501421@stu.ecnu.edu.cn}

\author{Chaoyi Yu}
\author{Yujie Yin}
\affiliation{%
  \institution{East China Normal University}
  \city{Shanghai}
  \country{China}}
\email{10235501470@stu.ecnu.edu.cn}
\email{10235501404@stu.ecnu.edu.cn}

\author{Zhongqian Mao}
\author{Jing Chen}
\affiliation{%
  \institution{East China Normal University}
  \city{Shanghai}
  \country{China}}
\email{10245101509@stu.ecnu.edu.cn}
\email{51285903016@stu.ecnu.edu.cn}

\author{Jiaqi Song}
\affiliation{%
  \institution{East China Normal University}
  \city{Shanghai}
  \country{China}}
\email{51285903003@stu.ecnu.edu.cn}

\author{Yunshi Lan}
\affiliation{%
  \institution{East China Normal University}
  \city{Shanghai}
  \country{China}}
\email{yslan@dase.ecnu.edu.cn}

\author{Yan Wang}
\authornote{Corresponding authors}
\affiliation{%
  \institution{East China Normal University}
  \city{Shanghai}
  \country{China}}
\email{yanwang@dase.ecnu.edu.cn}


\renewcommand\footnotetextcopyrightpermission[1]{}
\settopmatter{printacmref=false} 

\begin{teaserfigure}
  \centering
  \includegraphics[width=\textwidth]{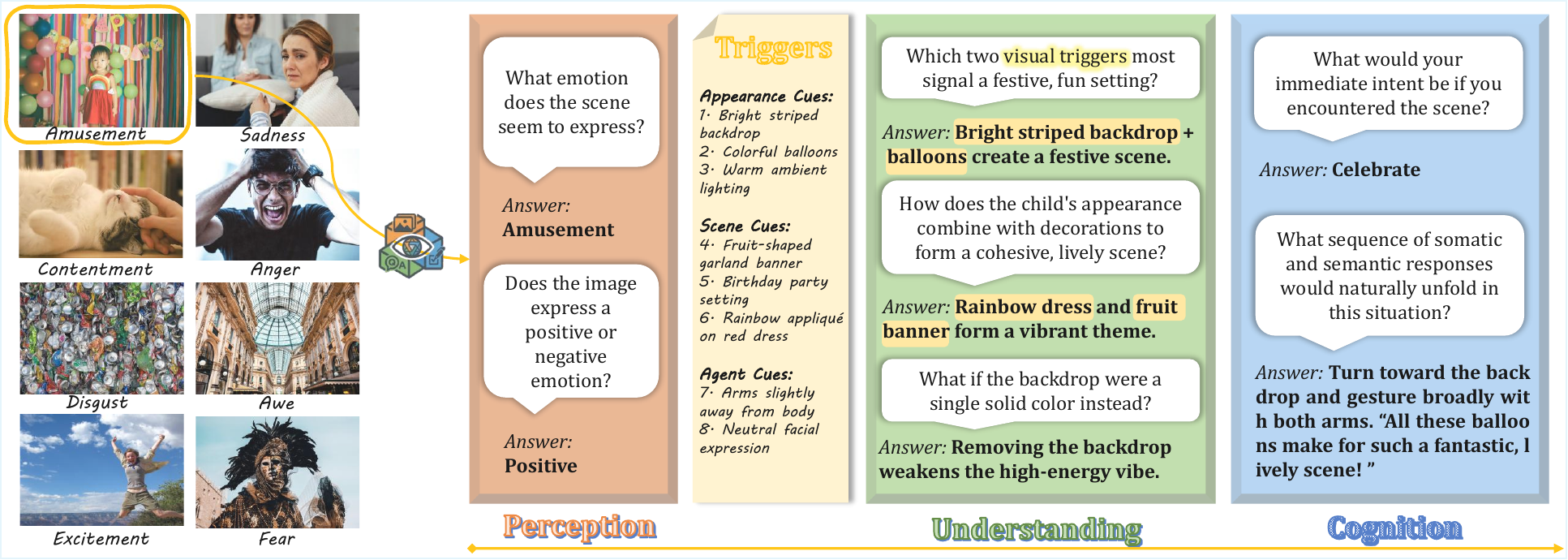}
  \caption{
  Overview of the InsightVQA Dataset. The proposed task decomposes multimodal emotional reasoning into three stages: Perception for recognizing emotion and valence, Understanding for grounding emotional triggers and Cognition for predicting response intent and performing sequential insight reasoning.}
  \label{fig:overview}
\end{teaserfigure}

\begin{abstract}

Visual emotion understanding requires models not only to recognize emotional states, but also to why they arise and perform higher-level cognitive reasoning. However, existing benchmarks mainly focus on emotion recognition, offering limited support for grounded understanding and response-oriented analysis. To address this gap, we introduce \textbf{InsightVQA}, a large-scale dataset for hierarchical visual question answering on emotion understanding and cognitive reasoning. Building from 351K images collected from six public sources, we apply a rigorous multi-stage filtering pipeline to curate 138K high-confidence images. Each image is annotated at three hierarchical levels: perception QA for emotion and valence recognition, grounded understanding QA constructed from visual trigger extraction through constraint-guided generation, and cognition QA centered on response intent prediction and sequential insight reasoning. In total, InsightVQA contains 725K QA pairs. We further present \textbf{InsightVQA-Bench}, a high-quality evaluation benchmark comprising 30K samples for fine-grained evaluation. To support evaluation, we introduce \textbf{InsightNet}, an emotion-tuned baseline for MLLMs. Results demonstrate that InsightVQA poses significant challenges for grounded emotion understanding and reasoning. The dataset will be publicly available at \url{https://akanthawang.github.io/InsightVQA}.

\end{abstract}





\maketitle

\section{Introduction}

A paradigm shift is underway in visual understanding, as multimodal large language models push the boundaries of artificial intelligence from surface-level perception to complex reasoning. Recent models demonstrate remarkable capabilities in integrating visual and textual information, enabling progress in tasks such as visual question answering, image captioning, and multimodal reasoning \cite{poria2017review, radford2021learning, alayrac2022flamingo, liu2023visual}. Beyond recognizing observable content, an emerging challenge is to infer human emotional and cognitive states from visual data, a capability that is crucial for affective human-computer interaction, socially assistive robotics, and emotion-aware support systems, and that requires deeper semantic interpretation and structured reasoning beyond mere label prediction\cite{kalateh2024systematic, andotra2023enhancing, wu2025comprehensive}.

Emotion understanding has traditionally been formulated as a classification or regression problem, focusing on predicting discrete emotion categories or continuous affective dimensions\cite{mollahosseini2017affectnet, kosti2017emotion}. While such approaches achieve strong performance on standard benchmarks, they largely capture correlations between visual features and affective labels, failing to account for contextual cues, situational factors, and observable triggers that shape real-world emotional responses. Recent advances in multimodal reasoning have motivated more sophisticated affective understanding; however, existing benchmarks still emphasize recognition over explanation, offering limited support for causal reasoning and cognitive interpretation\cite{haydarov2024affective, lin2025we, luo2026unveiling}. Consequently, current multimodal models struggle to identify emotion triggers, explain causal relationships, and generate coherent interpretations in complex scenarios, highlighting the need for structured benchmarks that enable interpretable and multi-level reasoning over emotional and contextual information.

Insights from cognitive science indicate that human understanding of others relies on the ability to infer latent mental states, including beliefs, intentions, and emotions. This capability, commonly referred to as Theory of Mind\cite{premack1978does, baron1985does}, enables individuals to reason about and predict the behavior of others based on limited observable cues. Subsequent research in social neuroscience and psychology further demonstrates that human social cognition is grounded in structured reasoning processes that integrate contextual information with affective and cognitive signals \cite{frith2006neural, adolphs2009social}. Consequently, effective computational modeling of human-centered visual understanding must move beyond low-dimensional label prediction. Instead, it demands representations that capture richer emotional and cognitive structures, providing the theoretical foundation for hierarchical and interpretable benchmarks that support multi-level reasoning.

Motivated by these considerations, we introduce \textbf{InsightVQA}, a large-scale dataset designed for hierarchical visual question answering that bridges emotion understanding and cognitive reasoning. Our proposed paradigm formulates human-centered visual understanding as a structured question answering task, enabling interpretable reasoning over emotional cues, contextual evidence, and inferred cognitive states. Unlike conventional visual question answering benchmarks that emphasize factual or descriptive content \cite{antol2015vqa, goyal2017making}, InsightVQA supports multi-level reasoning through three progressively deeper stages: perception, grounded understanding, and cognition.

A rigorous and quality-controlled annotation pipeline underpins the construction of the InsightVQA dataset. Starting with 351K images collected from six public sources, we conduct pseudo-label verification, class balancing, and expert validation to curate 138K high-quality images. Building on this foundation, we generate perception-level question-answer pairs to capture basic emotion and valence. For grounded understanding, we construct annotations through visual trigger extraction, trigger cleaning, constraint-guided generation, and iterative quality control. Finally, our cognition-level annotations extend the reasoning toward response intent and insight sequence analysis, derived from verified understanding evidence. The resulting dataset yields a total of 725K question-answer pairs.

To facilitate the standardized evaluation, we further present \textbf{InsightVQA-Bench}, a high-quality benchmark comprising 30K samples for fine-grained assessment. Using this benchmark, we evaluate representative multimodal large language models and introduce \textbf{InsightNet} as a strong baseline, trained with emotion instruction tuning and hierarchical preference optimization. Experimental results demonstrate that InsightVQA poses substantial challenges for grounded emotion understanding and cognitive reasoning, while enabling the systematic evaluation of advanced multimodal systems.

The main contributions of this work are summarized as follows:
\begin{itemize}[nosep,leftmargin=*]
    \item We introduce \textbf{InsightVQA}, a novel paradigm and extensive dataset comprising 138K curated images and 725K QA pairs for hierarchical visual question answering. Built through a rigorous annotation pipeline with strict quality control, it extends conventional emotion prediction toward interpretable and multilevel cognitive computing.

    \item We establish \textbf{InsightVQA-Bench}, a high-quality evaluation benchmark of 30K expert-validated samples, designed to support the fine-grained and standardized assessment of advanced multimodal reasoning capabilities.
    
    \item We develop \textbf{InsightNet} as a strong multimodal baseline. Extensive experiments demonstrate both the effectiveness of our model and the substantial challenges posed by the proposed benchmark.
\end{itemize}

\section{Related Work}

\begin{figure*}[!t]  
  \centering
  \includegraphics[width=\linewidth]{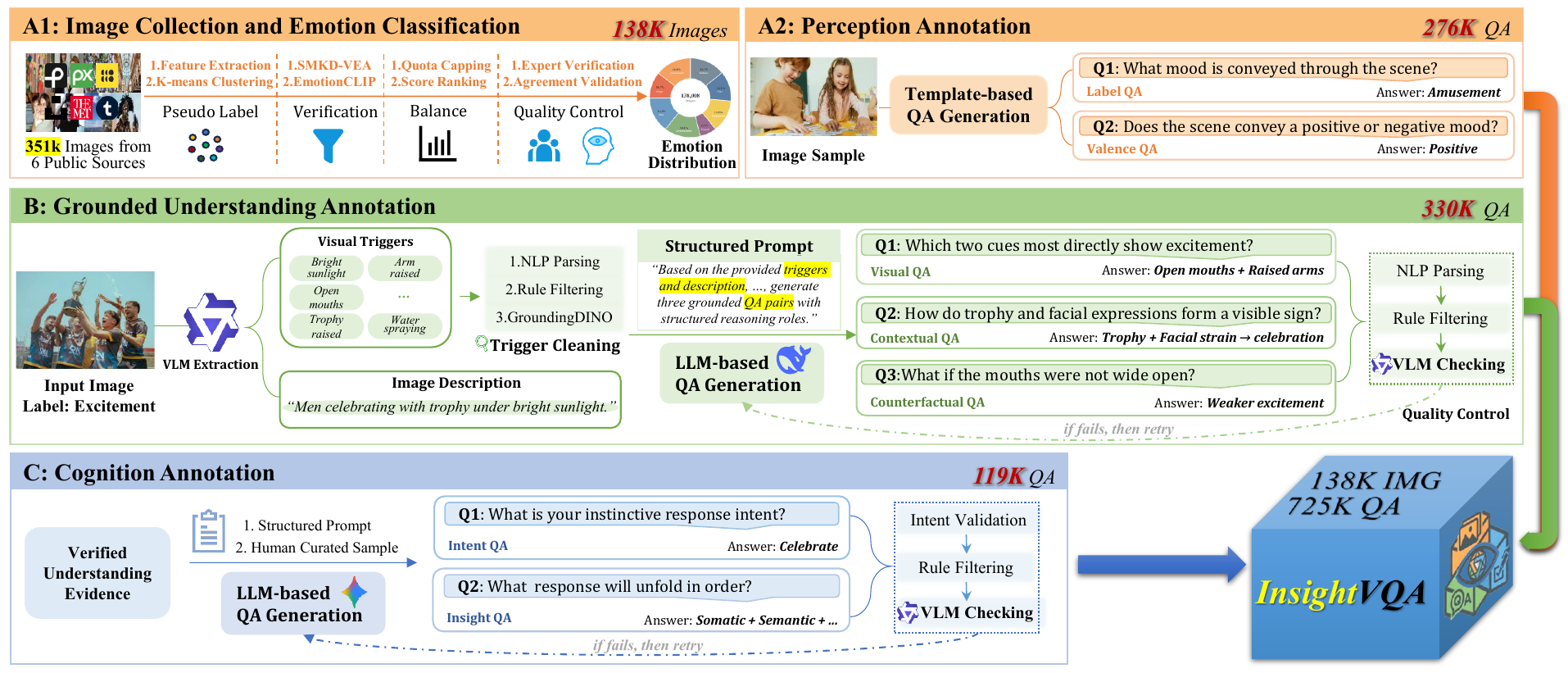} 
    \caption{
    InsightVQA dataset construction pipeline. 
    The pipeline comprises three stages: (A1) Image collection and emotion classification from 351K images across six public sources, producing 138K balanced samples; (A2) Perception annotation via template-based QA generation, yielding 276K label and valence QA pairs; (B) Grounded understanding annotation using VLM-extracted visual triggers and LLM-based QA generation, producing 330K visual, contextual, and counterfactual QA pairs. (C) Cognition annotation generates 119K intent and insight QA pairs via structured prompts and human curated samples, with quality controlled by rule filtering and VLM checking throughout. The final InsightVQA dataset comprises 138K images and 725K QA pairs.
  }
  \label{fig:dataset_pipeline}
\end{figure*}

Early datasets for image-evoked emotion understanding primarily focus on static affect annotation, providing either categorical emotion labels or dimensional scores. Representative datasets for categorical emotion recognition include Emotion6~\cite{peng2015emo6}, ArtPhoto~\cite{machajdik2010artphoto}, and EmotionROI~\cite{peng2016emotionroi}, while those based on dimensional affect modeling include EmoSet~\cite{yang2023emoset}. Larger-scale datasets such as FI~\cite{you2016fi} and EMOTIC~\cite{kosti2019emotic} incorporate richer contextual and social cues, enabling more realistic affective analysis.

Building on the development of visual question answering (VQA), recent datasets extend this paradigm to affective reasoning by pairing images with emotion-related questions or captions. EmoVIT \cite{xie2024emovit} frames emotion understanding as a VQA task, requiring models to identify emotional states from images and provide reasoning. EEmoDB \cite{gao2026eemo} provides structured QA pairs and evaluation sets to support learning and assessment of both basic emotion recognition and causal reasoning over emotional triggers. Meanwhile, benchmarks such as EmoBench-M \cite{hu2025emobenchm} , EEmo-Bench \cite{gao2025eemobench} and EmoBench \cite{yang2024emollm} offer standardized evaluation for affective tasks, measuring not only emotion recognition accuracy but also the quality of reasoning, explanation, and response generation in multimodal settings. 

However, existing datasets are primarily limited to coarse-grained emotion recognition, lacking explicit modeling of the underlying reasoning process and offering minimal supervision for capturing contextual dependencies and cognitive–affective interactions. As a result, they are insufficient for learning high-dimensional human state representations. To address this gap, we introduce InsightVQA, a large-scale dataset structured into perception, comprehension, and cognition stages to support multi-level reasoning, along with InsightVQA-Bench, a benchmark for evaluating contextual affective reasoning and cognitive–affective coupling.

\begin{table*}[t]
\centering
\caption{Comparison of affective VQA benchmarks in terms of scale, annotation, and task format. SCQ/MCQ/OEQ denote single-choice, multiple-choice, and open-ended questions; MTQ denotes multi-turn dialogue.}
\label{tab:dataset_comparison}
\small
\setlength{\tabcolsep}{5pt}
\begin{tabular}{l c c c c c c c c c}
\toprule
\textbf{Dataset} & \textbf{Year} & \textbf{\#Image} & \textbf{\#QA} & \textbf{Task Format} &
\makecell{\textbf{Human}\\\textbf{Assisted}} &
\makecell{\textbf{LLM}\\\textbf{Assisted}} &
\makecell{\textbf{Emotion}\\\textbf{Label}} &
\makecell{\textbf{Grounded}\\\textbf{Evidence}} &
\makecell{\textbf{Cognitive}\\\textbf{Reasoning}} \\
\midrule
EmoVIT~\cite{xie2024emovit}
& 2024 & 51.2K & 62.3K & SCQ, OEQ
& \cmark & \cmark & \cmark & \cmark & \xmark \\
AffectVisDial~\cite{haydarov2024affective}
& 2024 & 50K & 500K & MTQ, OEQ
& \cmark & \xmark & \cmark & \cmark & \xmark \\
MOSABench~\cite{song2024mosabench}
& 2024 & 1K & 1K & SCQ
& \cmark & \xmark & \cmark & \xmark & \xmark \\
EIBench~\cite{lin2025we}
& 2025 & 1.6K & 1.6K & OEQ
& \cmark & \cmark & \cmark & \cmark & \xmark \\
FaceBench~\cite{wang2025facebench}
& 2025 & 15.8K & 73.7K & SCQ, MCQ, OEQ
& \cmark & \cmark & \cmark & \xmark & \xmark \\
UNIFER~\cite{zhang2025rethinking}
& 2025 & 360K & 360K & SCQ
& \xmark & \cmark & \cmark & \cmark & \xmark \\
EEmoDB~\cite{gao2026eemo}
& 2026 & 128K & 1.2M & SCQ, OEQ
& \cmark & \cmark & \cmark & \cmark & \xmark \\
\rowcolor{lightblue}
InsightVQA (ours)
& 2026 & 138K(Original) & 725K & SCQ, MCQ, OEQ
& \cmark & \cmark & \cmark & \cmark & \cmark \\
\bottomrule
\end{tabular}
\end{table*}

\section{InsightVQA Dataset}
We introduce InsightVQA, a three-tier visual emotion QA dataset organized into perception, understanding, and cognition, as shown in \autoref{fig:dataset_pipeline}. This section presents the data construction pipeline, quality control process, and benchmark design of InsightVQA.

\subsection{Source Data and Scope} \label{sourcedata}
To construct a highly diverse and ecologically valid source pool for InsightVQA, we first curate 351,165 raw images from six major public repositories using a multi-dimensional heuristic search matrix. During our initial annotation analysis, we observed that general-purpose Vision-Language Models (VLMs) exhibit severe zero-shot affective biases—frequently defaulting to \textit{neutral} or \textit{contentment}. Consequently, we pivot to a specialized, high-fidelity tri-verification paradigm. We cross-validate unsupervised clustering pseudo-labels generated by a fine-tuned CLIP~\cite{radford2021clip} encoder against two state-of-the-art Visual Emotion Analysis (VEA) expert models, SMKD-VEA~\cite{cladiere2025skilldistill} and EmotionCLIP~\cite{zhang2023emotion}. To guarantee data purity, we strictly retain only those samples where at least one VEA model aligns with the clustering prediction, actively filtering out ambiguous and neutral-clustered images. Furthermore, to mitigate inherent real-world data distribution skews, such as the scarcity of the \textit{disgust} category, we implement a confidence-aware class balancing strategy, leveraging the averaged confidence scores of the VEA models to normalize majority classes. After purging the human-rejected samples, we finalize a well balanced, high-confidence perception foundation of 138,008 images, which provides a reliable basis for the subsequent understanding and cognition annotations.

\subsection{Hierarchical Dimension Annotation}

\noindent \textbf{Perception Annotation.}\label{3.2}
The perception layer serves as the entry-level annotation stage of InsightVQA by recasting categorical emotion annotations into a unified visual question answering format. For each image, two complementary question types are constructed: an eight-class emotion recognition task and a binary valence classification task. To mitigate template-level shortcut learning and enhance lexical diversity, each question type is instantiated with 20 linguistically varied templates spanning interrogative, conditional, and imperative framings. This design ensures that perception-level questions demand genuine visual understanding rather than superficial label lookup, while establishing a structurally consistent entry point that facilitates natural chaining into the higher-order reasoning required by subsequent layers.

\noindent \textbf{Understanding Annotation.} \label{3.3}
The understanding layer aims to elucidate \textit{why} an emotion is perceived by grounding the reasoning process in verifiable visual evidence. In our framework, a visual trigger is defined as an observable and image-grounded cue that provides direct evidence for inferring the depicted emotional state, rather than a speculative attribution of hidden mental states. To capture emotional evidence at different semantic granularities, we organize visual triggers into three complementary categories: low-level appearance cues, mid-level scene cues, and high-level agent cues. This categorization reflects the fact that emotion perception is typically supported not only by local perceptual details, but also by broader scene context and socially meaningful behaviors of the emotional subject.Specifically, we employ Qwen3-VL-32B-Instruct~\cite{bai2025qwen3} to extract candidate trigger phrases from each image across these three levels. The extracted triggers are subsequently validated through Grounding DINO~\cite{liu2024grounding} for spatial grounding and semantic filtering, thereby suppressing hallucinated or irrelevant candidates. Building upon the verified trigger set, DeepSeek-V3~\cite{liu2024deepseek} is used to generate three complementary QA types in a predefined sequential order: Visual Attribution, Contextual Synthesis, and Counterfactual Reasoning. Each QA instance is accompanied by an explicit reasoning statement grounded in the retained visual triggers. To ensure annotation fidelity, all generated samples are subjected to a cascaded quality control pipeline comprising rule-based validation, NLI-based consistency verification using DeBERTa-v3-large-mnli~\cite{he2021debertav3}, and visual re-examination via Qwen3-VL-32B-Instruct~\cite{bai2025qwen3}. Samples failing at any stage are routed to DeepSeek-V3~\cite{liu2024deepseek} for targeted revision and reintroduced into the pipeline for re-evaluation, forming an iterative process of generation, verification, and revision.

\noindent \textbf{Cognitive Annotation.}
The cognition layer formulates grounded response planning as a structured decision task, taking as input the outputs of the understanding layer rather than raw images. For each instance, an LLM is conditioned on few-shot exemplars retrieved from a manually curated exemplar bank based on the current evidence-grounded understanding results. It then generates a structured cognition annotation consisting of a response intent, an intent rationale, and a one-to-three-step \textit{insight\_sequence}. Each step in the \textit{insight\_sequence} is assigned to one of three categories, namely \textit{semantic}, \textit{somatic}, or \textit{regulatory}, and the annotation further includes explicit behavioral constraints and a safety note. To improve annotation reliability, the generated outputs are subjected to structural validation and an iterative VLM-based review process. Samples that are found to be inconsistent with the source image or the supporting reasoning evidence are automatically regenerated and re-evaluated.

\subsection{Human Verification}
In addition to the rigorous automated quality control procedures described above, we further perform human verification at two complementary levels to assess the reliability of InsightVQA. First, for the source images and their emotion labels, we conduct a large-scale manual inspection using stratified sampling proportional to the class distribution. A total of 6,400 images are reviewed by 8 expert annotators, with each image independently examined by 3 annotators. This process yields an overall pass rate of 92.52\% and a Fleiss' Kappa of 0.91, indicating strong inter-annotator agreement and low semantic ambiguity among the verified samples. Second, for the generated annotations, we conduct human verification on a stratified sample of 7,200 instances, sampled proportionally across the eight emotion categories and three annotation layers. To ensure broad coverage without redundancy, each of the three expert annotators independently evaluates a non-overlapping subset of 2,400 instances. The annotations are assessed on three criteria, yielding overall pass rates of 92.64\% in answer correctness, 95.00\% in visual grounding, and 94.58\% in reasoning rationality.

\begin{figure}[!t]  
  \centering
  \includegraphics[width=\linewidth]{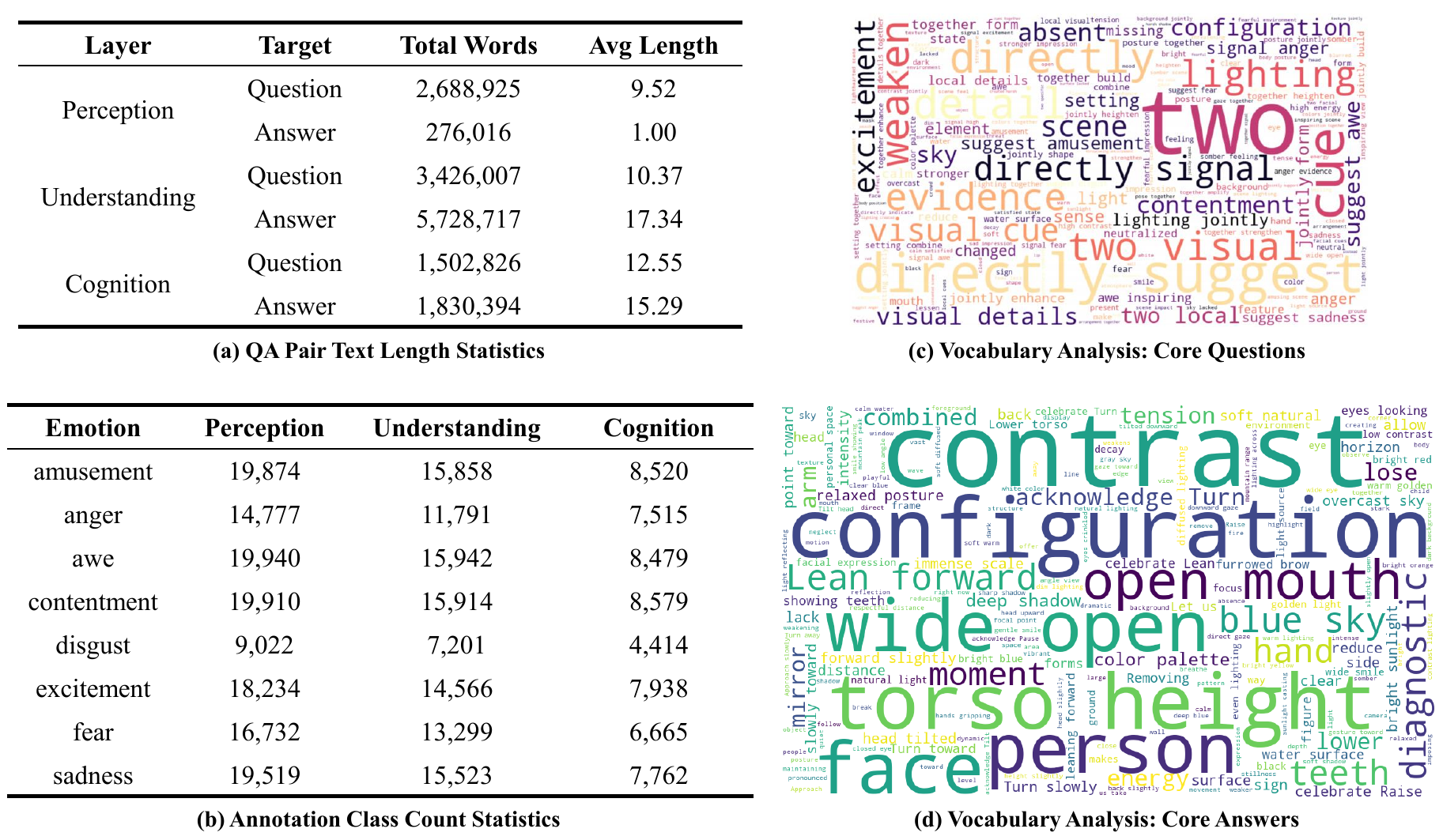} 
    \caption{
   Comprehensive dataset statistics and distributions of InsightVQA. 
  }
  \label{fig:data_statistics}
\end{figure}

\subsection{Benchmark Construction}
To support unified training and evaluation across InsightVQA, we randomly split the annotated data into 90\% training, retaining the generative QA format, and 10\% testing, standardized into discriminative tasks. The resulting benchmark comprises three hierarchical evaluation tasks. First, \textbf{Emotion Perception} assesses the recognition of emotion categories and valence through single-choice questions (SCQs). Second, \textbf{Emotion Understanding} evaluates evidence-grounded reasoning via SCQs covering visual attribution, contextual synthesis, and counterfactual reasoning. For both perception and understanding, each test instance couples the ground-truth answer with three semantically plausible, LLM-generated distractors. Finally, \textbf{Emotion Cognition} evaluates higher-order cognitive reasoning and insight generation. This task combines standard SCQs for identifying the most appropriate underlying intent with a Situational Judgment Test (SJT)~\cite{lievens2008situational} paradigm for evaluating concrete insight sequences. In the SJT phase, the model must rank multiple candidate insight sequences from best to worst based on the emotional state and supporting evidence, producing an ordered answer such as \textit{B, C, A, D}. This dual-format evaluation rigorously measures the model's ability to bridge emotional perception, intent formulation, and structured cognitive planning.

\subsection{Dataset Analysis and Statistics}

As shown in \autoref{fig:data_statistics}, the proposed InsightVQA dataset provides a large-scale, hierarchically structured collection of visual question-answering pairs that advances beyond flat label classification into a progressive cognitive framework—spanning \textit{Perception}, \textit{Understanding}, and \textit{Cognition}. Quantitative analysis validates this design: text length statistics (\autoref{fig:data_statistics}a) reveal a clear progression in textual complexity as cognitive depth increases, while the volume distribution (\autoref{fig:data_statistics}b) illustrates our rigorous filtering process across eight emotion categories. Furthermore, denoised vocabulary word clouds (\autoref{fig:data_statistics}c and d) demonstrate broad lexical diversity, highlighting a rich coverage of fine-grained visual features and complex contextual semantics. To contextualize these contributions, \autoref{tab:dataset_comparison} provides a comprehensive comparison with existing Affective VQA benchmarks, highlighting InsightVQA's unique inclusion of cognitive reasoning. Ultimately, InsightVQA captures diverse in-the-wild affective scenarios, providing a robust benchmark for evaluating high-dimensional cognitive-affective modeling.

\begin{table*}[htbp]
\centering
\caption{Comparison of Model Performance Across Multiple Tasks. }
\label{tab:performance}
\begin{tabular}{l c | c|c c c|c c c}
\toprule
\multirow{2}{*}{\textbf{Model}} & \multirow{2}{*}{\textbf{Year}} & \multicolumn{1}{c|}{\textbf{Perception}} & \multicolumn{3}{c|}{\textbf{Understanding}} & \multicolumn{3}{c}{\textbf{Cognition}} \\
\cmidrule{3-3} \cmidrule{4-6} \cmidrule{7-9}
 &  & ACC & F1 & Precision & Recall & Ranking & Top-1 & ACC \\
\midrule

\rowcolor{gray!15} \multicolumn{9}{l}{\textbf{Open-source MLLMs}} \\
\midrule
Qwen2.5-VL-7B \cite{bai2025qwen2.5vl} & 2025 & 57.95 &88.24  &87.74  &88.80  &34.95  &51.27  &30.92  \\
LLaVA-OneVision-1.5-8B \cite{xie025LLaVA-OneVision-1.5} & 2025 & 56.74 & 87.81 & 88.60 & 87.07 & 60.84 & 67.50 & 45.97 \\
InternVL3.5-8B \cite{wang2025internvl3} & 2025 & 53.40 &88.54  &88.53  &88.60  &31.81  &43.44  &31.18  \\
Deepseek-VL-7B-chat \cite{lu2024deepseek} & 2024 &52.54  &87.25  &88.42  &86.12  &40.52  &43.48  &38.07  \\
Qwen2.5-VL-32B \cite{bai2025qwen2.5vl} & 2025 &52.80  & 88.04  &87.90  &88.19  &69.24  &65.32  &44.02  \\
Qwen2.5-VL-72B \cite{bai2025qwen2.5vl} & 2025 &53.83  &88.86  &88.13  &89.62  &76.30  &56.83  &57.24  \\

\midrule
\rowcolor{gray!15} \multicolumn{9}{l}{\textbf{Close-source MLLMs}} \\
\midrule
Qwen3-Max~\cite{bai2025qwen3}  & 2026 & 52.64 & 87.73 & 86.96 & 88.53 & 75.65 & 61.14 & 42.51 \\
Deepseek-V3.2~\cite{liu2025deepseek}  & 2025 & 51.20 & 87.86 &87.23 & 88.53 & 80.24 & 65.77 & 41.60 \\
GPT-4o \cite{hurst2024gpt} & 2024 & 50.63 & 88.34 & 88.49 & 88.24 & 77.51 & 61.06 & 50.34 \\
Gemini-2.5-flash \cite{comanici2025gemini} & 2025 &56.83  &89.05  &88.02  &90.13  &80.56  &63.51  &56.08  \\
Claude-3.7-sonnet \cite{kasireddy2026evaluating} & 2026 &56.37  &88.72  &89.16  &88.31  &81.77  &63.49  &61.87  \\

\midrule
\rowcolor{gray!15} \multicolumn{9}{l}{\textbf{Emotion-Oriented MLLMs}} \\
\midrule
EmoViT~\cite{xie2024emovit} & 2024 & 53.69 & 84.50 &84.85   &84.19  & $-$  &12.53  &42.86  \\
Emotion-Qwen~\cite{huang2025emotion} & 2025 & 52.98  &86.92  & 86.37  &87.48  & $-$ & 33.61 & 30.18  \\
EmoCaliber~\cite{wu2025emocaliber} & 2025 & 40.29 &84.23  &81.52  &87.18  & $-$ & 25.47  & 27.92 \\
\rowcolor{lightblue} InsightNet (Ours) & 2026 & \textbf{76.25} & \textbf{90.56} &\textbf{90.50} & \textbf{90.63} &\textbf{82.79}  &\textbf{71.21}  &\textbf{69.18}  \\
\bottomrule
\noalign{\smallskip} 
\multicolumn{9}{l}{\footnotesize "$-$" indicates that the model is not applicable to the evaluation.} \\
\end{tabular}
\end{table*}

\section{InsightNet}

\begin{figure}[h]  
  \centering
  \includegraphics[width=\linewidth]{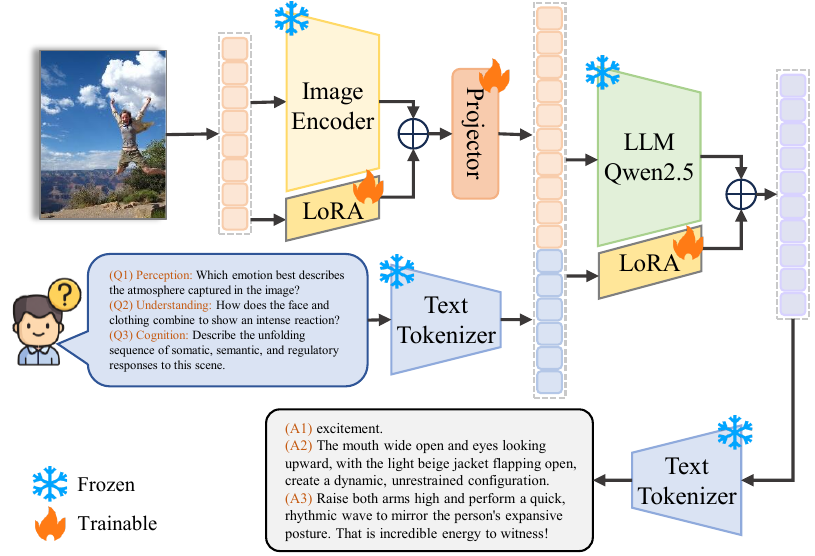} 
    \caption{
    Training framework of InsightNet. We perform supervised LoRA-based fine-tuning on InsightVQA to equip the model with foundational question-answering capabilities.
  }
  \label{fig:framework}
\end{figure}

Built upon the InsightVQA dataset, we propose InsightNet, a unified architecture for human state understanding, as shown \autoref{fig:framework}. To instill foundational reasoning capabilities, we perform LoRA-based fine-tuning on InsightVQA, enabling the model to capture hierarchical perception, comprehension, and cognition for high-dimensional cognitive–affective representations.

To enhance both cognitive depth and generalization in complex emotional scenarios, we further apply instruction-driven supervised fine-tuning (SFT) on Qwen2.5-VL-7B using the InsightVQA dataset.
As full-parameter fine-tuning is prone to overfitting on emotionally biased data \cite{kumar2022sft}, we adopt LoRA \cite{hu2022lora} as an implicit regularization mechanism, reducing trainable parameters via low-rank decomposition.

For a pre-trained weight matrix $W_0 \in \mathbb{R}^{d \times k}$, the adapted weight is formulated as:

\begin{equation}
    h = W_0 x + \Delta W x = W_0 x + B A x
\end{equation}
\noindent where $\Delta W = BA$, with $A \in \mathbb{R}^{r \times k}$ and $B \in \mathbb{R}^{d \times r}$ being trainable low-rank matrices, and $r \ll \min(d, k)$.

The training objective minimizes the Negative Log-Likelihood (NLL) over the InsightVQA dataset $\mathcal{D}$:

\begin{equation}
\mathcal{L}_{\text{SFT}} = 
\mathbb{E}_{(I, Q, Y) \sim \mathcal{D}} 
\left[ - \sum_{t=1}^{T} 
\log P(y_t \mid I, Q, y_{<t}; \Theta \cup \{A, B\}) \right]
\end{equation}
\noindent where $I$ denotes the input visual tensor, $Q$ represents the affective cognition instruction, and $\Theta$ refers to the frozen pre-trained parameters. Through this approach, the model learns to transform implicit internal affective associations into explicit textual representations.

\section{Experiments}
\subsection{Settings of Experiments}
\noindent \textbf{Implementation Details.} We implement the InsightNet training pipeline in PyTorch, utilizing Qwen2.5-VL-7B as the foundational vision-language model. The model is fine-tuned on the InsightVQA dataset using Low-Rank Adaptation (LoRA) to ensure parameter efficiency. Our LoRA configuration consists of a rank $r = 32$, a scaling factor $\alpha = 32$ and a dropout rate of 0.05. During training, we employ the AdamW optimizer with a cosine learning rate scheduler to achieve stable convergence.

\noindent \textbf{Evaluation Metrics.} We evaluate model performance using multiple metrics: accuracy (ACC) for classification tasks; precision, recall and F1-score for open-ended tasks; ranking-based metrics such as Top-1 accuracy and the Spearman rank correlation for ranking tasks.

\subsection{Quantitative Results}
As shown in \autoref{tab:performance}, InsightNet achieves state-of-the-art performance across the three tasks in InsightVQA-Bench. We elaborate on these results as follows:

\noindent\textbf{Perception Task.} Evaluated by accuracy, InsightNet consistently outperforms baseline models across various emotional categories. This superiority suggests that our instruction tuning effectively aligns visual representations with fine-grained affective labels, enabling the model to capture subtle emotional cues that are often overlooked by general-purpose MLLMs.

\noindent\textbf{Understanding Task.} To assess open-ended response quality, we employ BERTScore to measure semantic similarity between model outputs and ground-truth references. As illustrated in \autoref{tab:performance}, our model achieves a significantly higher BERTScore than competing methods. This indicates that InsightNet generates more contextually relevant and linguistically precise descriptions of the emotional drivers within complex social scenes.

\noindent\textbf{Cognitive Task.} In the Situational Judgment Test (SJT) paradigm, InsightNet demonstrates a sophisticated ability to rank candidate insight sequences. The high Spearman Rank Correlation between the model’s predicted rankings and the ground-truth sequences confirms that the model transcends basic perception. It effectively evaluates the logical consistency of behavioral insights, reflecting a high-order capability for structured cognitive planning and intent formulation.

The consistent performance gains across these tasks highlight a clear synergy: accurate emotional Perception provides the necessary grounding for deeper Understanding, which in turn facilitates the complex Cognitive Reasoning required for the SJT task. The hierarchical improvement validates the effectiveness of our multi-layered instruction data.

\section{Conclusion}

In this work, we introduce InsightVQA, a hierarchical visual emotion question answering benchmark designed to support fine-grained emotion understanding across three progressive levels: Perception, Understanding, and Cognition. This tiered structure enables systematic evaluation of emotion recognition, contextual reasoning, and decision-oriented response generation. To facilitate comprehensive assessment, we construct a large-scale dataset comprising approximately 725K question-answer pairs alongside a curated evaluation set for multi-level benchmarking. Extensive experiments conducted on both open-source and closed-source multimodal large language models demonstrate that, while existing models attain relatively strong performance on lower-level perception tasks, they continue to exhibit substantial limitations in higher-level emotion understanding and cognitive reasoning. These findings highlight the need for more capable models that integrate perceptual grounding with deeper affective and contextual reasoning. We hope InsightVQA will serve as a rigorous and valuable resource for advancing hierarchical affective understanding and grounded multimodal reasoning, providing a solid foundation for future research on context-aware visual emotion understanding and emotionally intelligent multimodal systems.

\newpage


\bibliographystyle{ACM-Reference-Format}
\bibliography{sample-base}

\clearpage

\appendix

\section{Detailed Data Construction and Quality Assurance Pipeline }\label{Appendix A}

To guarantee the transparency and reproducibility of the InsightVQA dataset, this section details the comprehensive data collection, the empirical findings on VLM bias, and the rigorous class-balancing procedures introduced in Section \ref{sourcedata}. 

\subsection{Data Collection and Unsupervised Pseudo-Labeling }

We systematically crawl images from six prominent platforms (e.g., Pexels, Pixabay, Metmusem) using a search matrix spanning five macro-dimensions (Action, Facial, Style, Entity, Concept) paired with over 1,500 fine-grained emotional keywords. After exact and near-duplicate removal, we establish an initial pool of 351,165 unannotated images. To generate heuristic baseline labels, we fine-tune a CLIP vision encoder on the EmoSet dataset to extract emotion-discriminative 768-dimensional features. K-Means clustering ($K=9$) is applied, and the resulting clusters are manually mapped to eight target emotions and one "neutral" category. 
\begin{figure}[h]  
\centering  
\includegraphics[width=\linewidth]{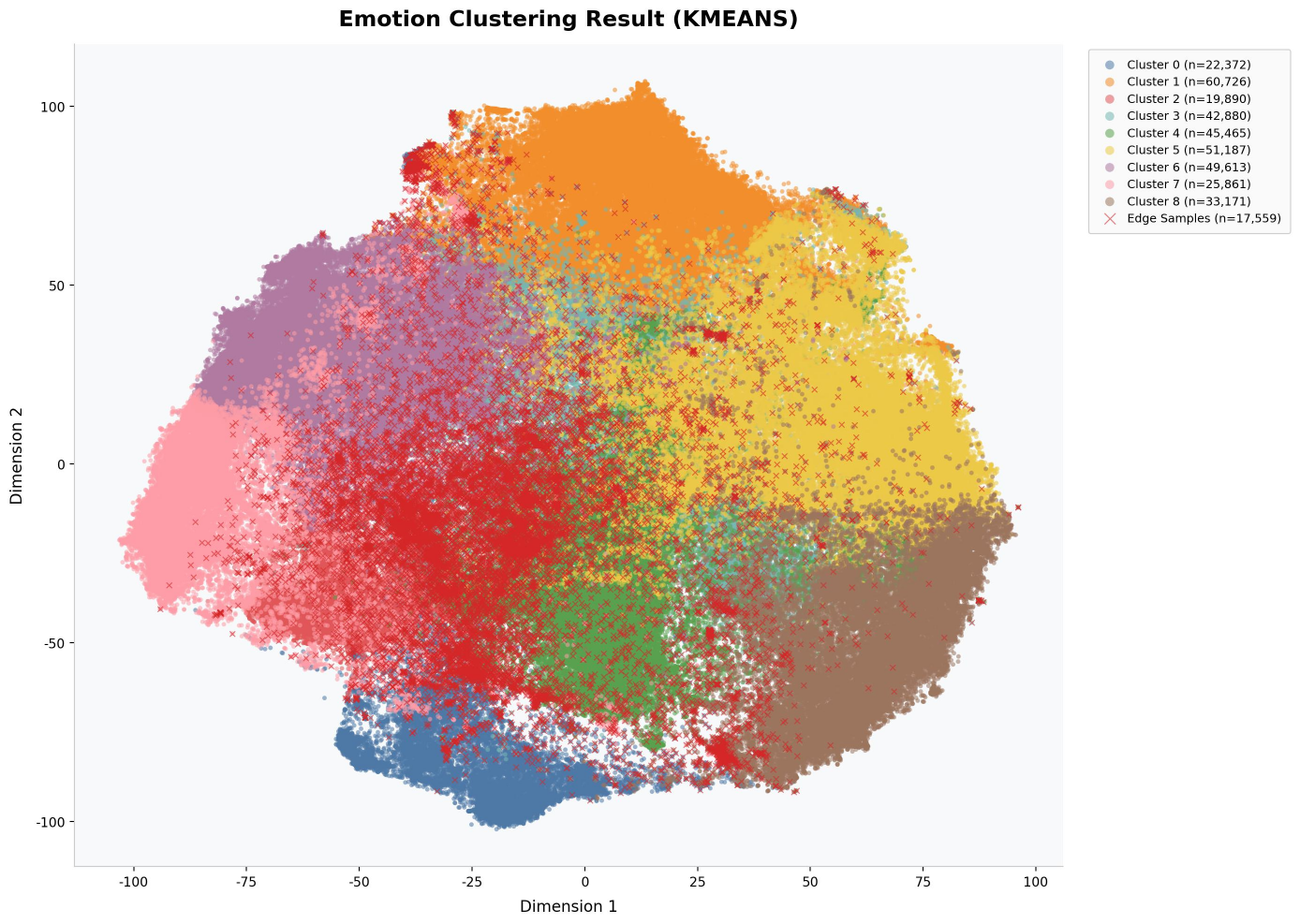}  
\caption{    
\textbf{t-SNE Visualization of Unsupervised Emotion Clustering.}
    We apply K-Means ($K{=}9$) on fine-tuned CLIP features extracted from 351,165 images.
    Each point represents an image, colored by its assigned cluster.
    Edge samples (marked with $\times$) are subsequently discarded.
  }  
\label{fig:cluster_tsne}
\end{figure}
\subsection{Empirical Analysis of VLM Affective Bias }

\begin{figure}[h]  
\centering  
\includegraphics[width=\linewidth]{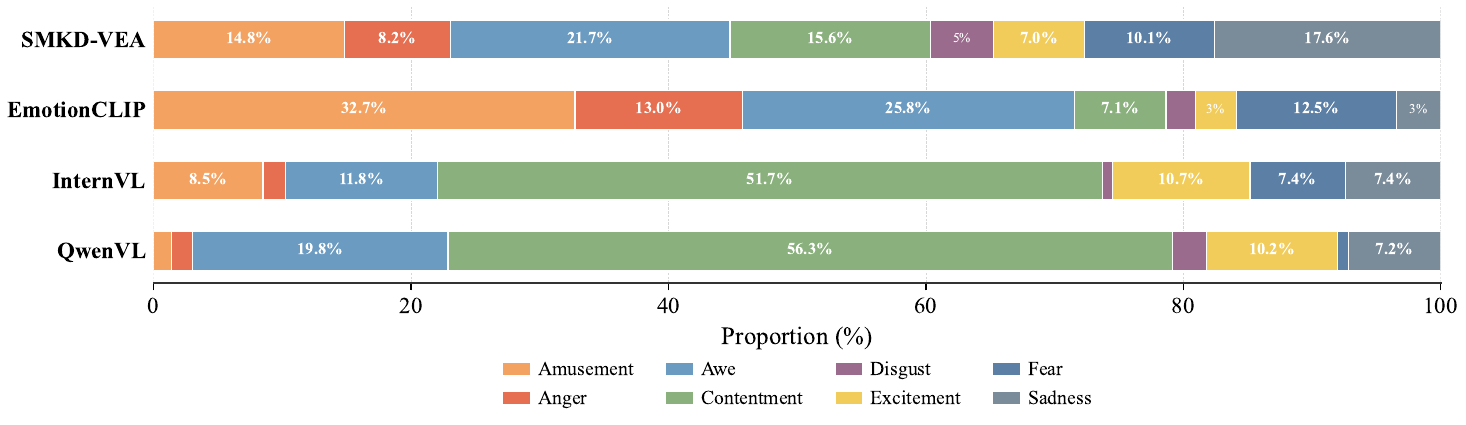}  
\caption{    
\textbf{Emotion Label Distribution of Four Models.}
    SMKD-VEA and EmotionCLIP produce relatively balanced distributions,
    while InternVL and QwenVL exhibit severe affective bias,
    with over 50\% of predictions concentrated on ``contentment''.
  }  
\label{fig:vlm_bias}
\end{figure}

Initially, we attempted a five-model consensus voting mechanism incorporating advanced VLMs, including QwenVL and InternVL. However, our empirical distribution analysis (see \autoref{fig:vlm_bias}) revealed profound affective biases in general-purpose VLMs. Specifically, QwenVL~\cite{bai2023qwen} and InternVL~\cite{chen2024internvl} skewed heavily positive, categorizing over 50\% of the dataset as "contentment". Due to this severe zero-shot bias, which compromises the fine-grained emotional diversity required for embodied reasoning, we entirely excluded these VLMs from the perception annotation pipeline. 

\subsection{Tri-Verification and Strict Filtering }
To ensure annotation fidelity, we employ a specialized tri-verification system comprising the clustering pseudo-labels and two dedicated Visual Emotion Analysis models (SMKD-VEA and EmotionCLIP). The filtering proceeds as follows:
\begin{enumerate}
    \item We first discard all samples assigned to the "neutral" cluster, as well as marginal edge samples identified during the t-SNE visualization.
    \item A strict retention policy is applied: an image is retained if and only if the prediction of \textit{at least one} VEA model matches the clustering pseudo-label. All other conflicting instances are outright discarded to prevent noise propagation. This stage distills the dataset down to 214,015 high-confidence images.
\end{enumerate}

 \begin{figure}[h]
  \centering
  \includegraphics[width=\linewidth]{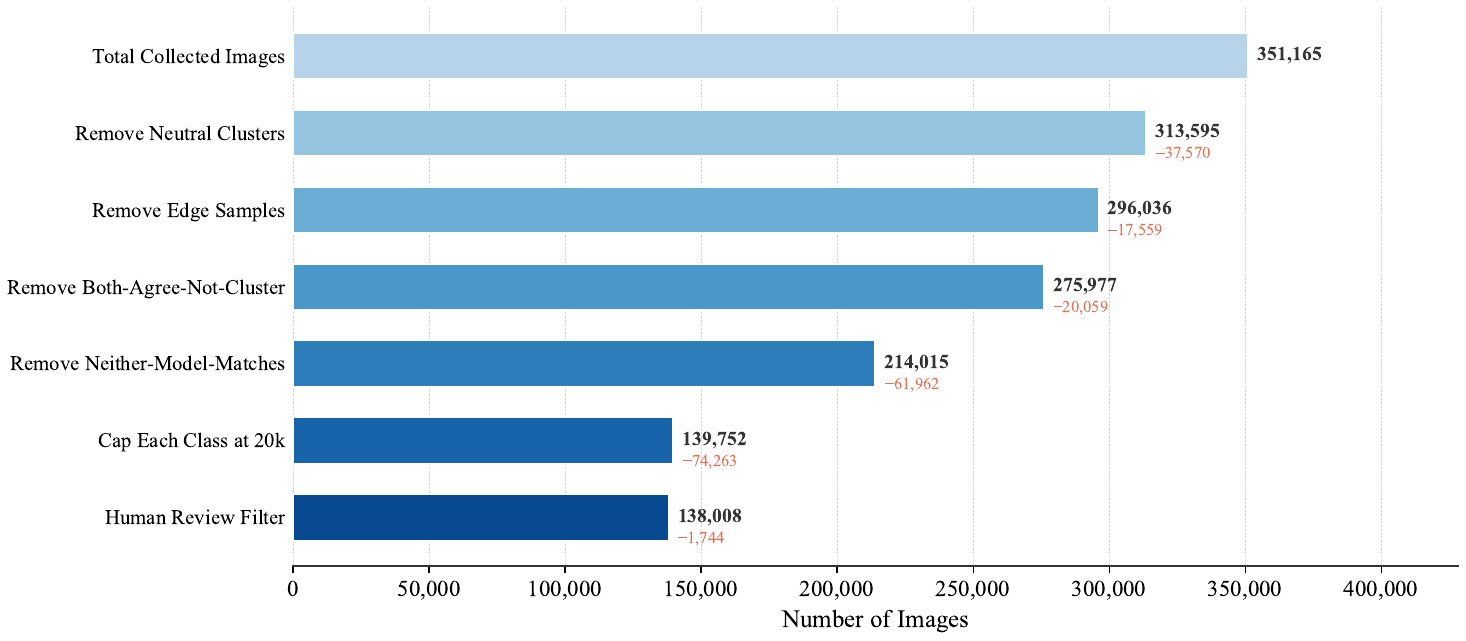}
  \caption{
    \textbf{Data Filtering Pipeline Funnel.}
    Starting from 351,165 collected images, each stage progressively
    removes low-confidence samples. The red segments indicate
    the number of images discarded at each step, yielding a final
    set of 138,008 verified images after human review.
  }
  \label{fig:pipeline_funnel}
\end{figure}

\begin{figure}[t]
  \centering
  \includegraphics[width=\linewidth]{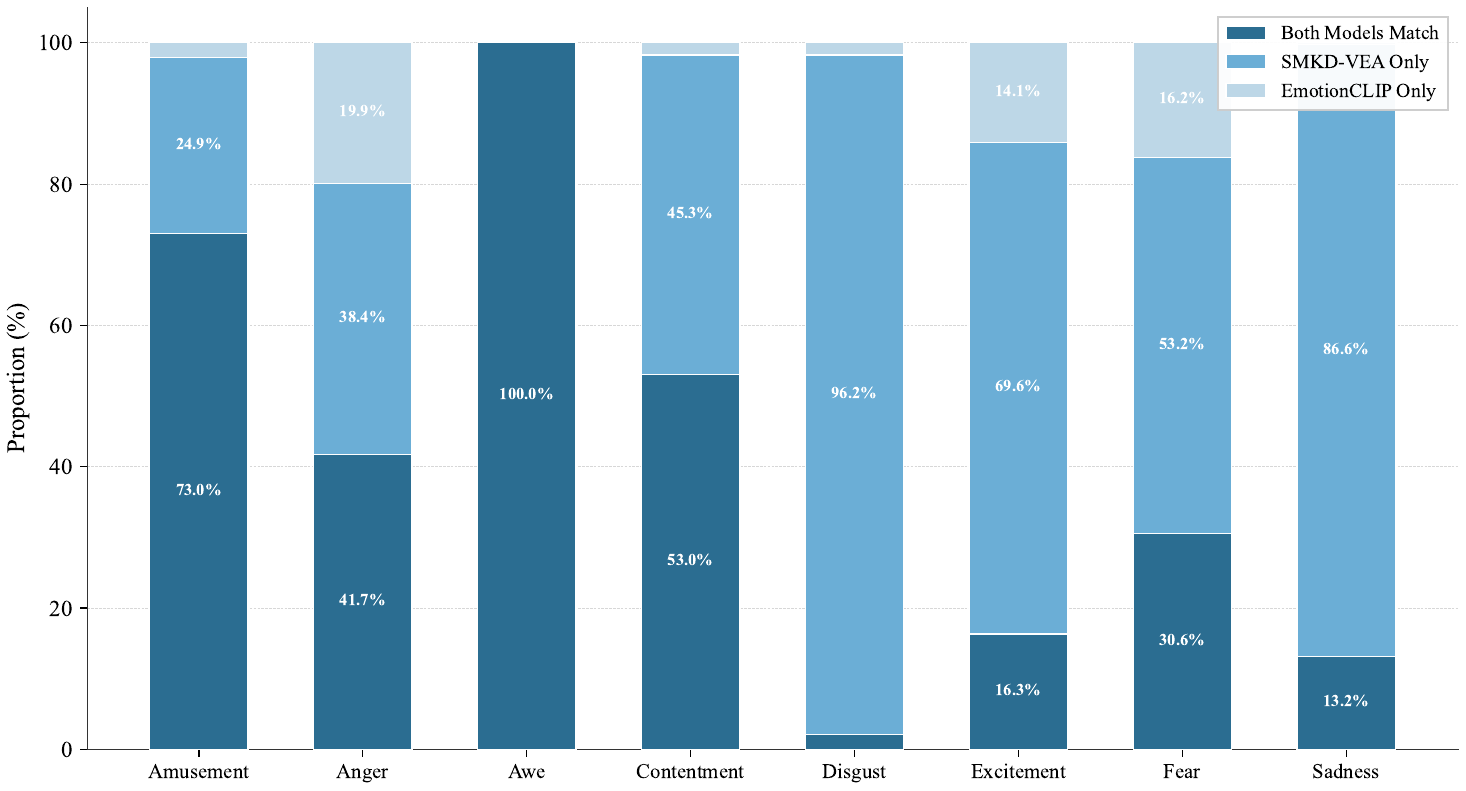}
  \caption{
    \textbf{Tri-Verification Tier Breakdown per Emotion.}
    For each category, we show the proportion of images retained
    by full three-way consensus (both VEA models match clustering),
    SMKD-VEA-only match, and EmotionCLIP-only match.
    Categories like ``awe'' achieve near-perfect three-way agreement,
    while ``disgust'' relies predominantly on single-model verification.
  }
  \label{fig:tier_breakdown}
\end{figure}

\subsection{Confidence-Aware Class Balancing }
Post-filtering, a severe class imbalance was observed (e.g., the "disgust" category contained only 9,206 valid images). To prevent the InsightNet baseline from developing long-tail bias, we cap the maximum number of images per category at 20,000. For over-represented categories, we prioritize samples that achieved perfect three-way consensus (Clustering == SMKD-VEA == EmotionCLIP). To fill the remaining quota up to 20,000, we rank the remaining two-way consensus samples based on the average prediction confidence scores of the two VEA models in descending order. This strategic sampling yields a highly balanced intermediate dataset of 139,752 images. 

\begin{figure}[h]  
\centering  
\includegraphics[width=\linewidth]{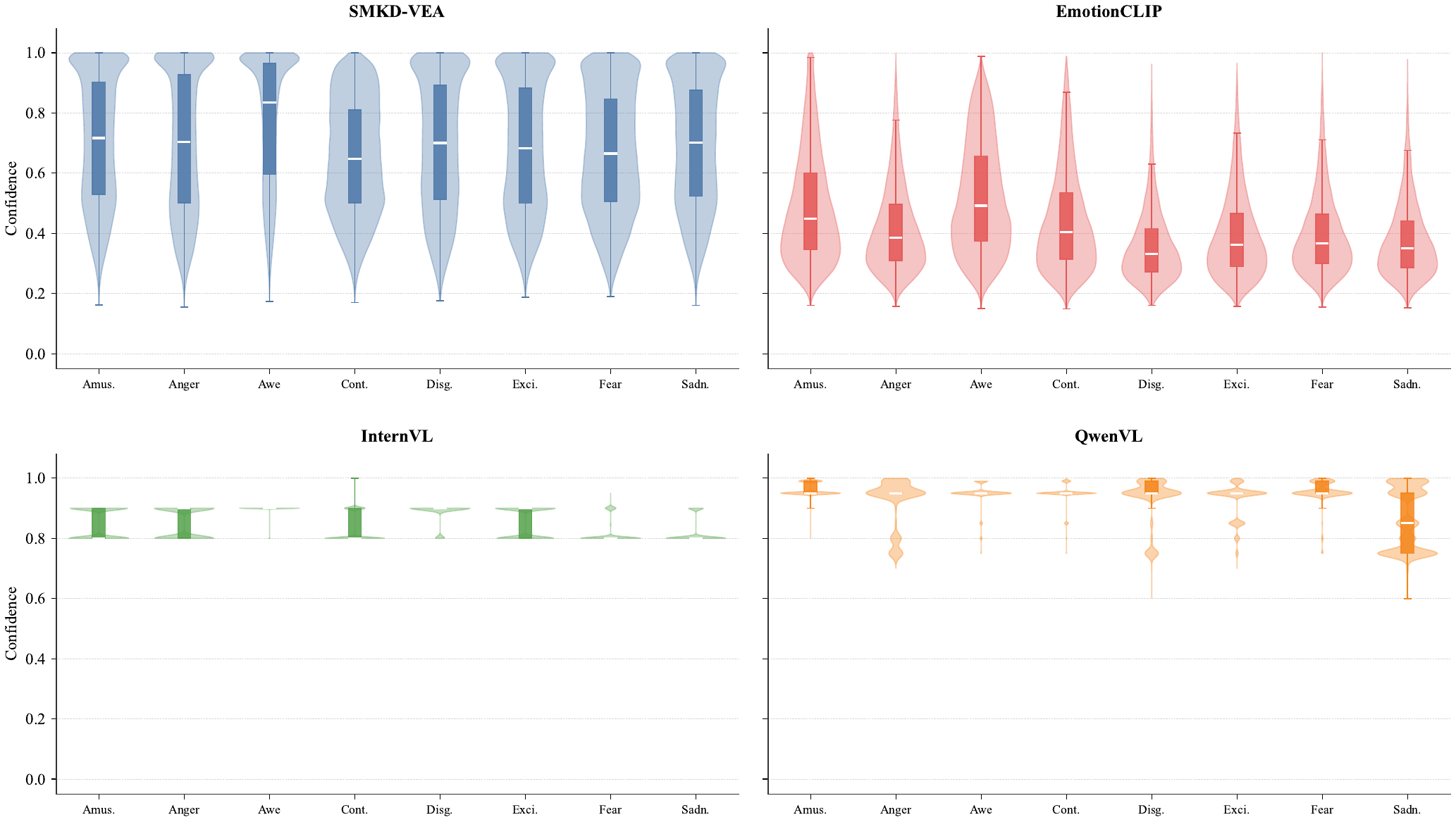}  
\caption{    
\textbf{Confidence Score Distributions of Four Models.}
    Box plots of per-category prediction confidence for SMKD-VEA,
    EmotionCLIP, InternVL, and QwenVL.
    The confidence-aware balancing strategy prioritizes high-confidence
    samples when capping over-represented categories at 20,000.
  }  
\label{fig:confidence_dist}
\end{figure}

\subsection{Rigorous Human Quality Inspection }
To quantitatively validate the dataset's reliability, we conduct a large-scale human verification phase. We apply stratified sampling proportional to the class distribution to select 6,400 images. These images are distributed among 8 expert annotators. To ensure robust inter-rater reliability, the workload is mathematically structured such that every single image is independently annotated by exactly 3 individuals ($6,400 \times 3 = 19,200$ total tasks). Consequently, each annotator evaluates exactly 2,400 images ($19,200 \div 8 = 2,400$), with an perfectly even distribution across all emotional categories.
The inter-annotator agreement yields a remarkable PassRate of 0.925, demonstrating exceptional dataset quality and minimal semantic ambiguity. Finally, any image rejected during this human verification phase is permanently purged from the dataset. The finalized perception layer of InsightVQA comprises 138,008 pristine, class-balanced images, providing a robust empirical foundation for complex affective understanding. 

\begin{figure}[h]  
\centering  
\includegraphics[width=\linewidth]{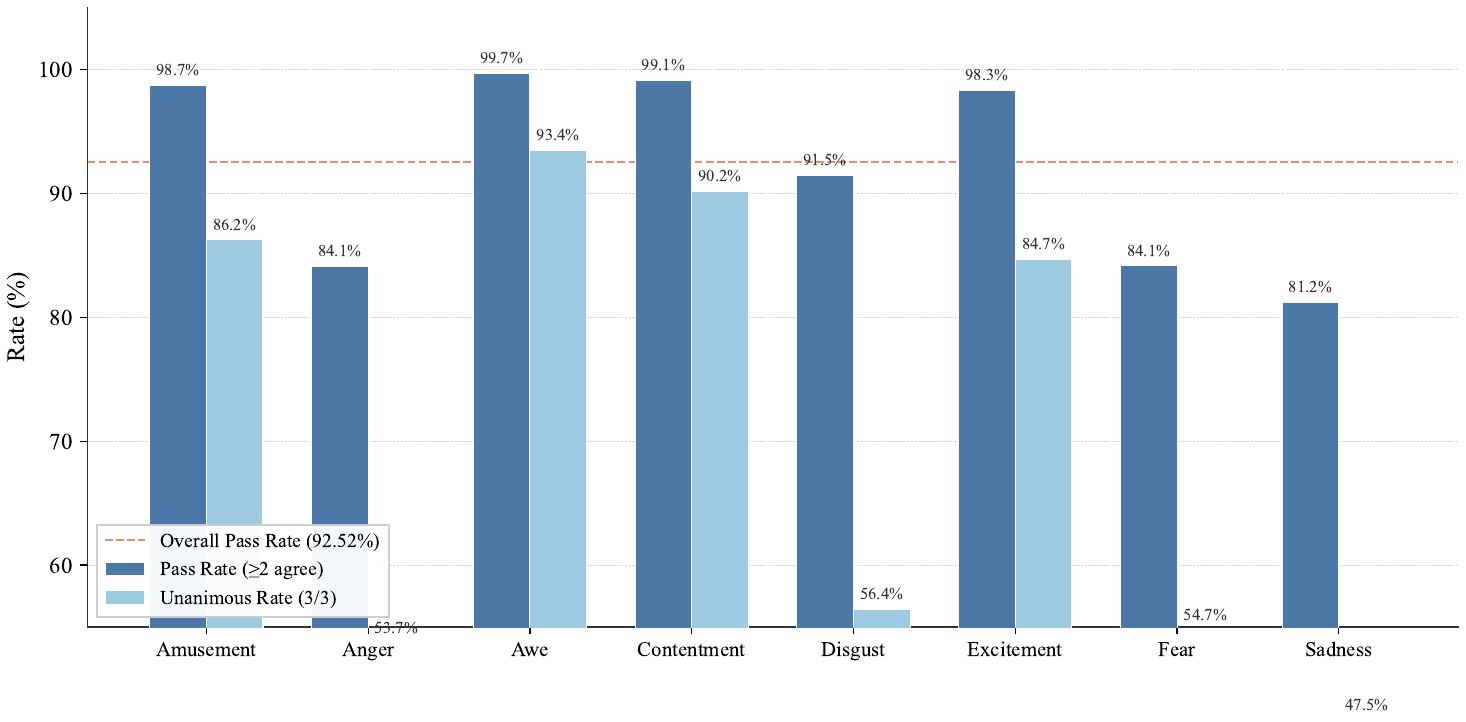}  
\caption{    
\textbf{Human Verification Results per Emotion Category.}
    Blue bars indicate the pass rate ($\geq$2 of 3 annotators agree),
    and orange bars indicate the unanimous agreement rate (3/3).
    The overall pass rate is 92.52\%, with a Fleiss' Kappa of 0.91.
  }  
\label{fig:human_review}
\end{figure}

\begin{figure}[h]  
\centering  
\includegraphics[width=0.95\linewidth]{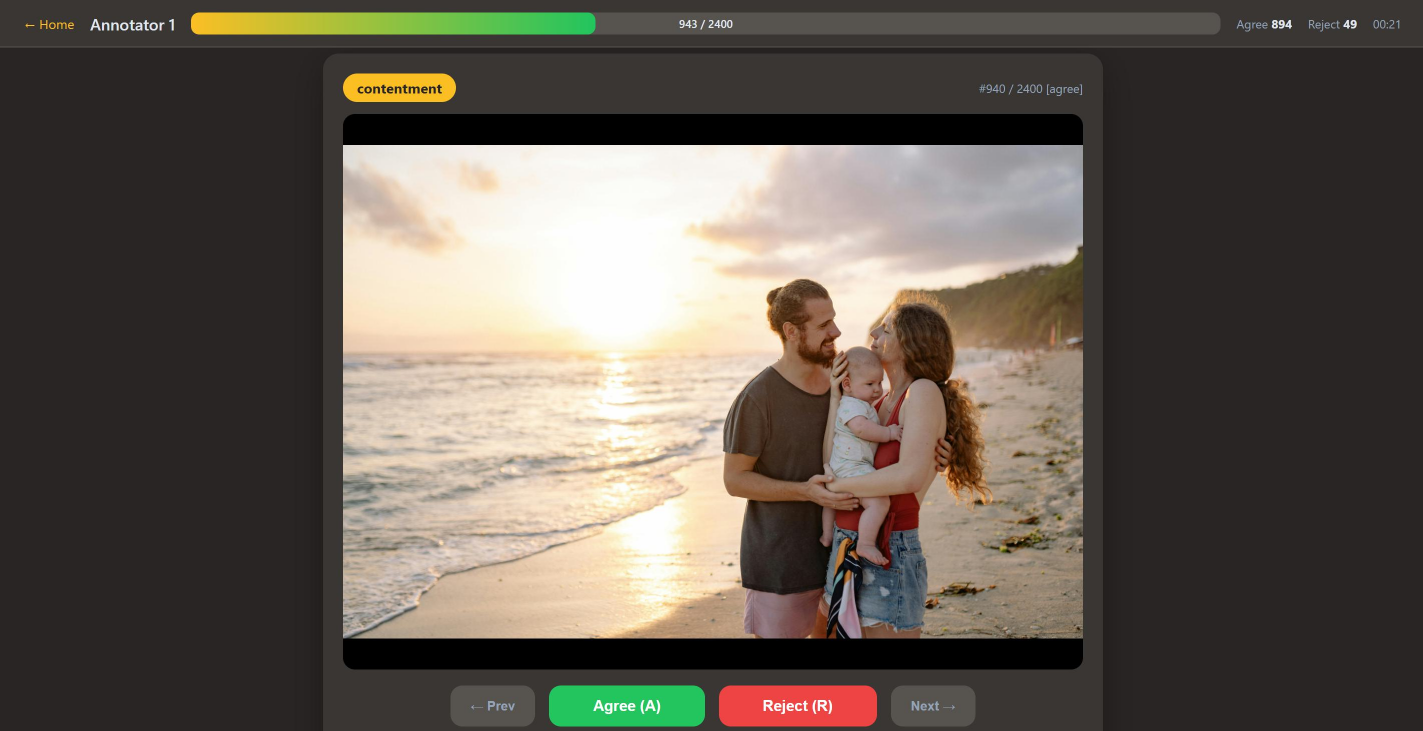}  
\caption{    
\textbf{Web-Based Human Annotation Interface.}
    Each annotator reviews images via a browser-based tool with
    real-time progress tracking, keyboard shortcuts, and instant
    server-side persistence. Annotators can agree with the assigned
    label or reject with a corrected emotion category.
  }  
\label{fig:annotation_tool}
\end{figure}

\begin{figure}[h]  
\centering  
\includegraphics[width=0.85\linewidth]{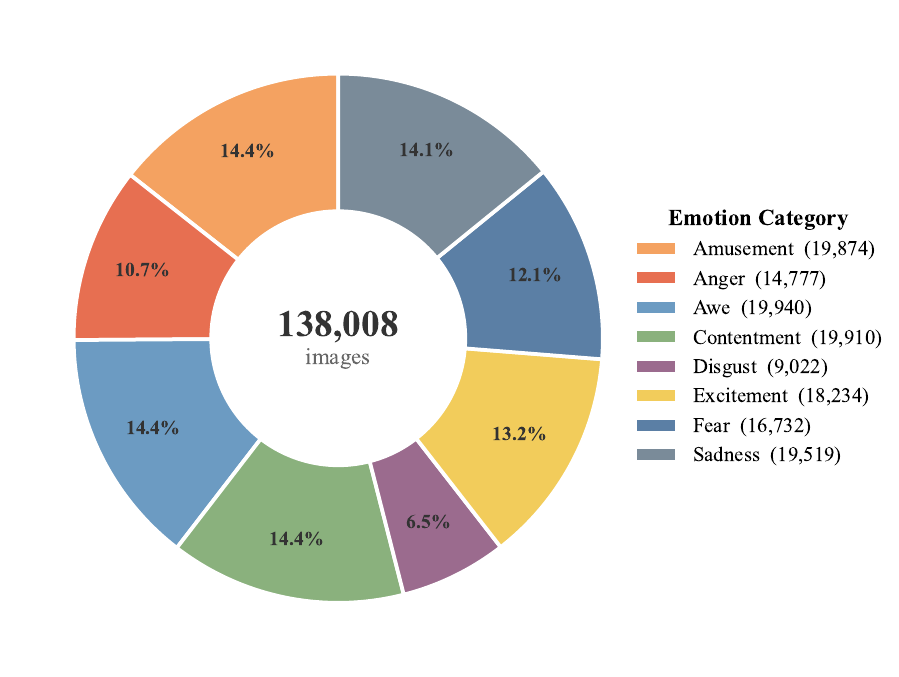}  
\caption{    
\textbf{Final InsightVQA Perception Layer Distribution.}
    After tri-verification, class balancing, and human review filtering,
    the finalized dataset contains 138,008 images across eight
    emotion categories with near-uniform distribution.
  }  
\label{fig:final_distribution}
\end{figure}

\section{Detailed Annotation Pipeline}
\label{app:pipeline}

\begin{figure*}[t]
\centering
\includegraphics[width=\textwidth]{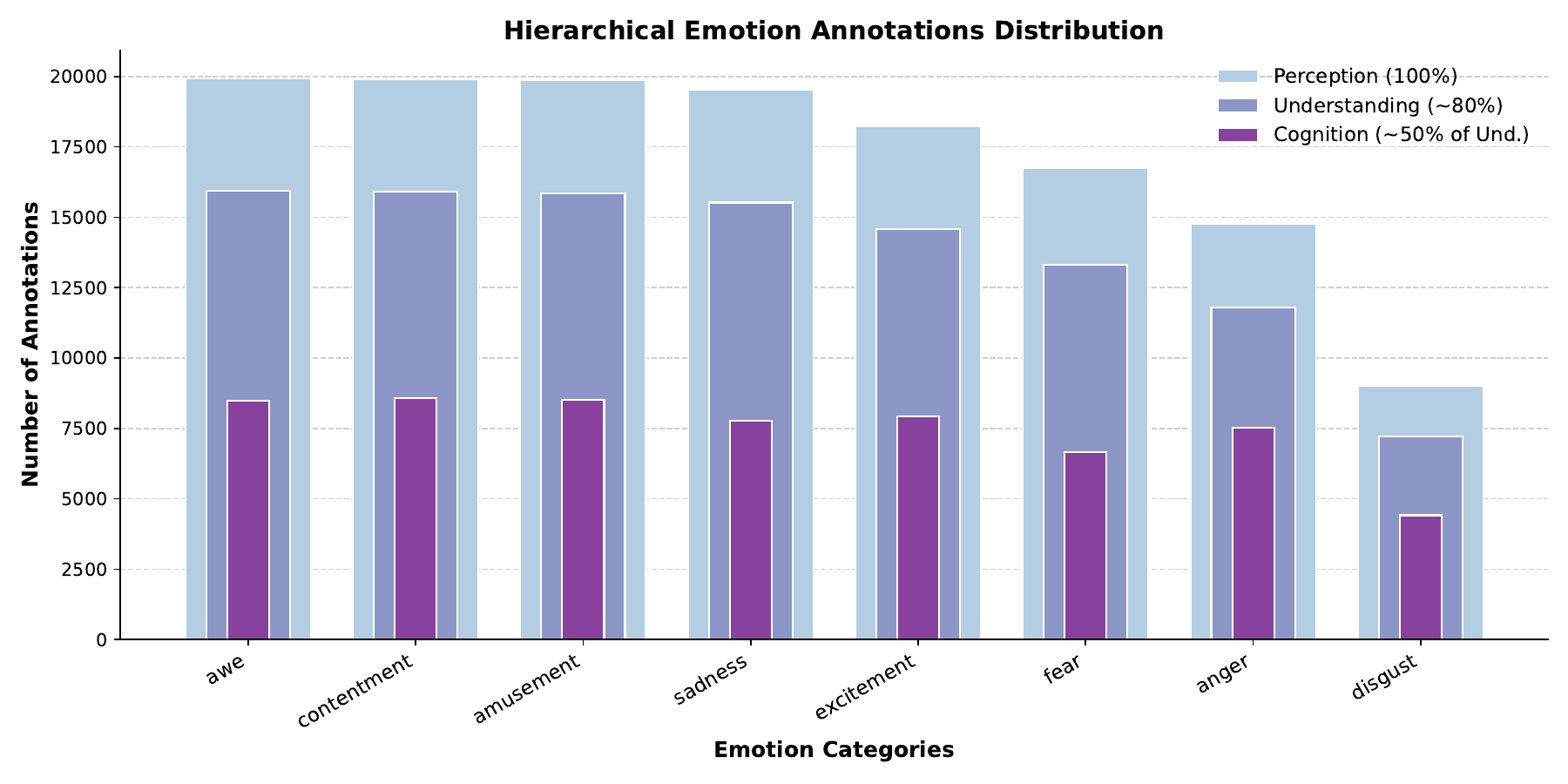}
\caption{Distribution of Hierarchical Emotion Annotations.
}
\label{fig:hierarchical_anno_distribution}
\end{figure*}

This appendix provides the complete technical specification of the hierarchical annotation pipeline introduced in Section~\ref{3.2}. An overview of the annotation distribution across the three layers is shown in \autoref{fig:hierarchical_anno_distribution}.

\subsection{Perception Layer Details}

\begin{table*}[t]
\centering
\caption{Emotion Questions with Labels and Valence.}
\label{tab:qa_templates}
\begin{tabular}{p{3cm} p{10cm}}
\toprule
\textbf{Category} & \textbf{Question} \\
\midrule
\multirow{20}{*}{Valence} 
 & Is the emotional tone of the image positive or negative? \\
 & Does the scene convey a mostly positive or negative mood? \\
 & Would you describe the emotion in the image as positive or negative? \\
 & Is the overall feeling in the image more positive or more negative? \\
 & Does the emotional atmosphere lean toward positivity or negativity? \\
 & Is the general emotional impact positive or negative? \\
 & Does the image express a broadly positive or negative emotion? \\
 & Would the emotional tone be classified as positive or negative? \\
 & Is the emotion shown more positive or negative in nature? \\
 & Does the image communicate a positive or negative emotional state? \\
 & Should the overall emotion be labeled as positive or negative? \\
 & Is the mood presented here best described as positive or negative? \\
 & Does the emotional quality of the image feel positive or negative? \\
 & Is the dominant emotional impression positive or negative? \\
 & Does the picture give off a positive or negative emotional vibe? \\
 & Is the displayed emotion positive or negative overall? \\
 & Would you categorize the emotion as positive or negative? \\
 & Is the emotional expression in the scene positive or negative? \\
 & Does the scene create a positive or negative emotional impression? \\
 & Is the emotional response evoked by the image positive or negative? \\
\midrule
\multirow{20}{*}{Label} 
 & What emotion does the scene seem to express? \\
 & What feeling is reflected in the overall atmosphere of the image? \\
 & What emotional state best describes the moment shown? \\
 & What kind of emotion does the subject appear to be experiencing? \\
 & What mood is conveyed through the scene? \\
 & What emotional tone dominates the image? \\
 & How would you describe the emotional impression the image gives? \\
 & What emotion is suggested by the expressions or context? \\
 & What is the primary feeling represented in the image? \\
 & What emotion best matches the visual cues in the scene? \\
 & What is the dominant emotional state depicted here? \\
 & What kind of feeling does the image evoke? \\
 & How does the emotion in the image appear to be expressed? \\
 & What emotional reaction might be inferred from the scene? \\
 & What emotion seems most appropriate to describe the image? \\
 & What type of emotion can be observed in the moment shown? \\
 & What emotional message does the scene communicate? \\
 & What is the likely emotion conveyed by the visual elements? \\
 & What feeling is most strongly suggested by the scene? \\
 & Which emotion best describes the atmosphere captured in the image? \\
\bottomrule
\end{tabular}
\end{table*}

For the perception layer, we formulate two classification tasks as QA pairs: emotion recognition over eight categories and binary valence prediction. To promote linguistic diversity and reduce superficial pattern matching, we manually design 20 question templates for each task, yielding 40 templates in total, as shown in \autoref{tab:qa_templates}. The templates vary in wording and sentence structure while preserving the same underlying query intent. For each image, the emotion label is read from the perception-stage manifest, and the valence label is deterministically derived from the emotion category via a predefined positive-negative mapping. During data construction, one template is sampled at random from each set, producing one emotion QA pair and one valence QA pair per image. Both labels are inherited from the tri-verified perception foundation described in Section~\ref{sourcedata}.

\subsection{Understanding Layer Details}

\begin{figure}[!h]
    \centering
    \includegraphics[width=\linewidth]{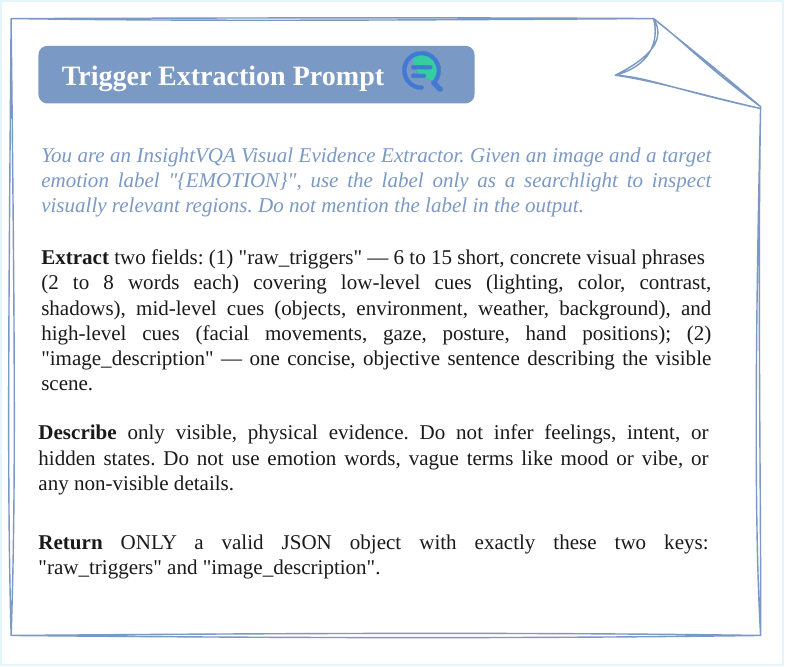}
    \caption{Trigger Extraction Prompt.}
    \label{fig:trigger_extract}
\end{figure}

\noindent\textbf{Stage 1: Trigger Extraction.}
For the understanding layer, we use Qwen3-VL-32B-Instruct~\cite{bai2025qwen3} to extract a set of short, objective visual triggers from each image. The model is prompted with the target emotion as a searchlight for visual inspection rather than as output text, and returns a JSON object containing two fields: raw triggers 
and image description. The former consists of 6 to 15 short phrases, spanning low-level appearance cues (illumination, color temperature, contrast), mid-level scene elements (objects, weather, spatial layout), and high-level bodily signals (facial movements, gaze direction, posture). The latter is constrained to a single objective sentence describing the overall scene. To prevent label leakage, the prompt explicitly excludes emotion words, mental-state terms, causal explanations, and any non-visible details. Inference is run at temperature 0.1 with max tokens=512. The full prompt templates used for trigger extraction is shown in \autoref{fig:trigger_extract}.

\noindent\textbf{Stage 2: Trigger Verification.}
The extracted triggers are refined through a two-step verification process designed to suppress hallucination and reduce redundancy. We first apply lightweight linguistic filtering with spaCy~\cite{vasiliev2020natural} to remove overly generic, repetitive, or weakly informative phrases, while extracting noun-chunk candidates for subsequent grounding. Spatially localizable triggers are then validated against the image using Grounding DINO~\cite{liu2024grounding} ($\tau_{\text{box}}=0.30$, $\tau_{\text{text}}=0.25$), with images resized to a maximum edge of 768 pixels and processed in FP16. If a full phrase fails grounding, the system retries with its root noun. To avoid false negatives on inherently global or non-localizable evidence, phrases involving facial parts, gaze, lighting, contrast, background, or texture bypass box-based rejection. The surviving triggers form the verified set, which serves as the evidence source for subsequent annotation stages.

\begin{figure}[!h]
    \centering
    \includegraphics[width=\linewidth]{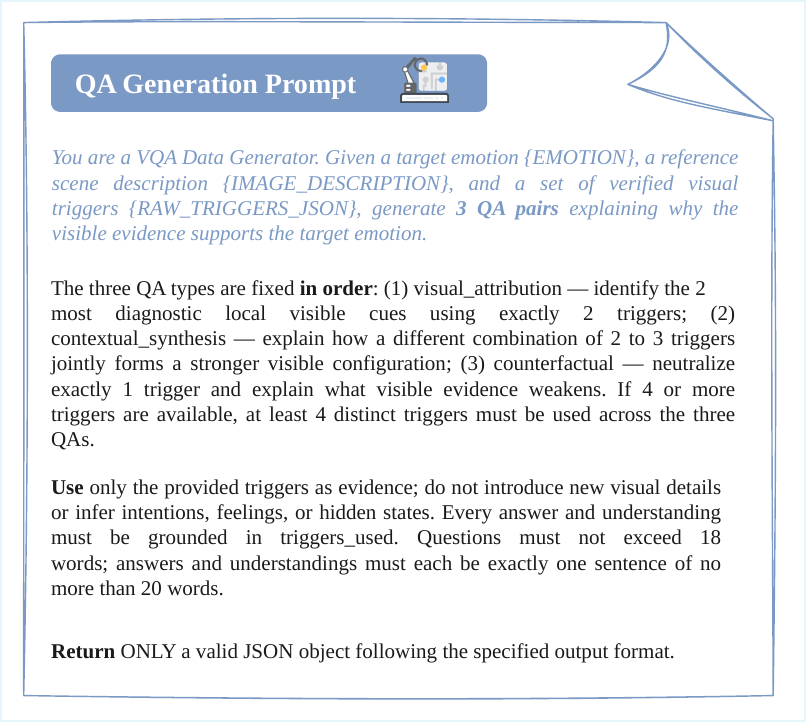}
    \caption{QA Generation Prompt.}
    \label{fig:qa_generation}
\end{figure}

\noindent\textbf{Stage 3: QA Generation.}
Given the verified trigger set and the factual image description, we employ the large language model DeepSeek-V3.2~\cite{liu2024deepseek} to generate exactly three question-answer (QA) pairs per image in JSON format. The full prompt templates is provided in \autoref{fig:qa_generation}. During this process, the image description serves purely as reference context for overall scene understanding, while the verified triggers act as the strict, sole evidence source for answer generation. The three generated QA pairs correspond to visual attribution, contextual synthesis, and counterfactual reasoning. These question types are designed to assess different levels of affective understanding: from identifying salient local cues, to integrating multiple triggers into a coherent explanation, and finally to reasoning about how the visible emotional evidence would weaken if a key cue were neutralized. Each QA pair is recorded with structured identifiers (e.g., id, type, and difficulty) and contains four core fields: question, answer, understanding, and used triggers. Additionally, the output file preserves metadata such as the image ID, emotion label, model name, and evidence path. To guarantee evidence diversity and annotation faithfulness, we enforce explicit trigger-allocation constraints across the three pairs: if four or more verified triggers are available for an image, the generated answers must encompass at least four distinct triggers in total. Furthermore, to prevent hallucination, no answer is permitted to introduce visual details beyond those explicitly listed in its associated used triggers.

\begin{figure}[!h]
    \centering
    \includegraphics[width=\linewidth]{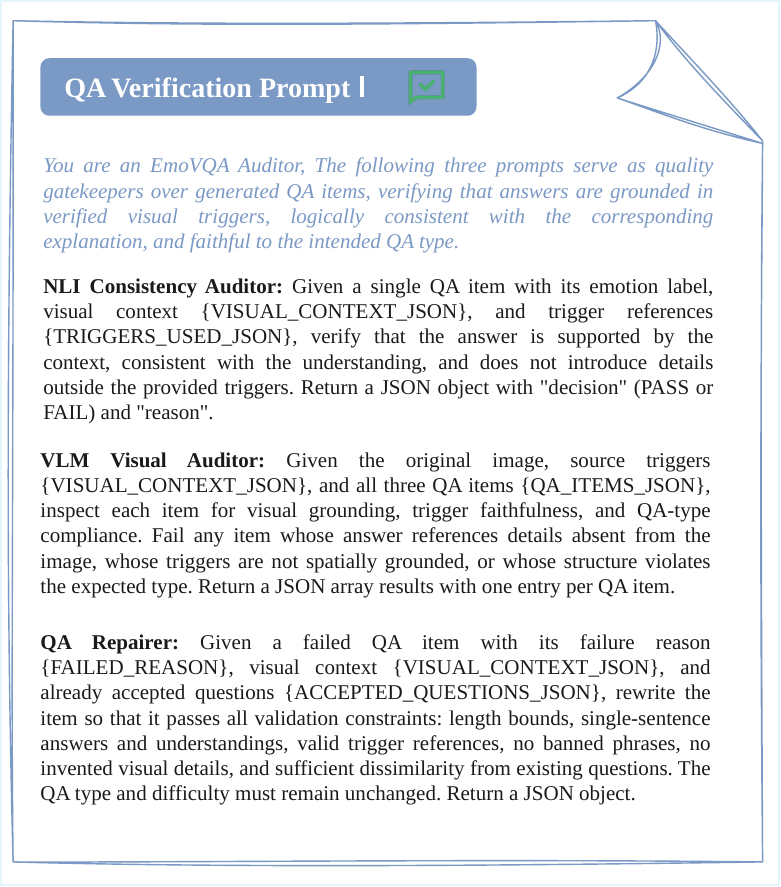}
    \caption{QA Verification Prompts.}
    \label{fig:qa_verification}
\end{figure}

\noindent\textbf{Stage 4: Quality Control Funnel.}
Each generated sample passes through a three-stage cascaded pipeline before being accepted into the dataset. The rule-based stage enforces word-count bounds on all textual fields, restricts answers and explanations to a single sentence, and validates trigger references against the verified trigger set, capping each QA pair at three triggers and requiring exactly one for counterfactual items. It also rejects abstract or uncertainty-expressing phrasing via a banned-expression list and enforces cross-question diversity using Jaccard similarity ($\theta{=}0.8$). Next, DeBERTa-v3-large fine-tuned on MNLI~\cite{he2021debertav3} checks logical consistency between each answer and its explanation, rejecting high-confidence contradictions ($p_{\text{contradiction}}{>}0.9$). Finally, Qwen3-VL-32B~\cite{bai2025qwen3} re-examines the original image with the full QA triple, verifying that cited triggers are visually grounded, answers are faithful to the referenced evidence, and each pair conforms to its intended question type. The full prompt templates is provided in \autoref{fig:qa_verification}. Samples failing any stage are sent to DeepSeek-V3~\cite{liu2024deepseek} for targeted revision and re-evaluated in the same pipeline, with up to three repair iterations per sample.

\subsection{Cognition Layer Details}

\begin{figure}[!h]
    \centering
    \includegraphics[width=\linewidth]{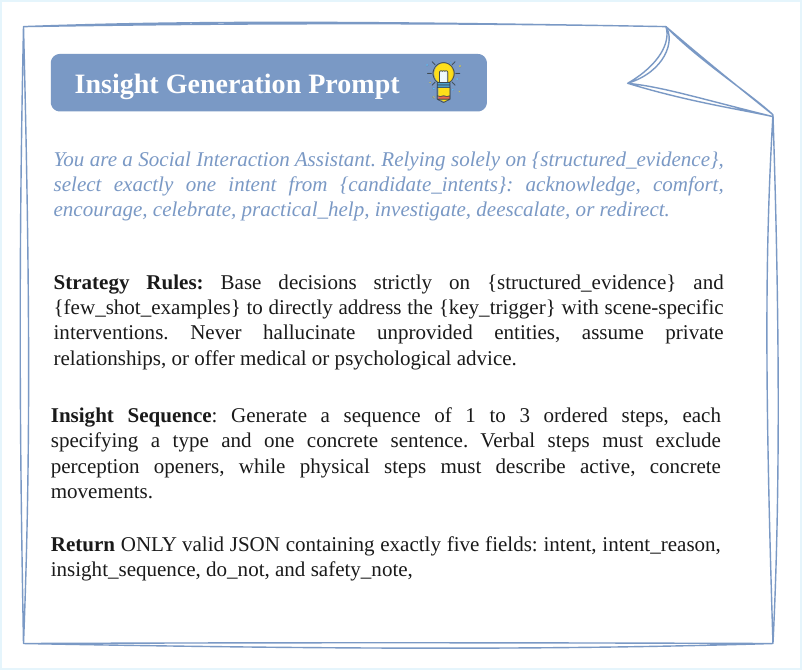}
    \caption{Insight Generation Prompt.}
    \label{fig:insight_generation}
\end{figure}

\noindent\textbf{Insight Generation.}
The cognition layer transforms the verified outputs of the understanding layer into a structured insight response. To support this process, we define a closed set of eight response intents, namely acknowledge, comfort, encourage, celebrate, practical help, investigate, deescalate, and redirect. For each image, the planner first constructs a compact evidence representation from the understanding layer by retaining at most two non-counterfactual question-answer pairs, with priority given to visual attribution and then contextual synthesis, while the trigger associated with the counterfactual question is separately preserved as the key trigger. Response generation is further supported by a pool of 24 exemplars written by experts, with three exemplars provided for each emotion category. Retrieval first favors exemplars from the same emotion class and only falls back to other emotions when necessary. Candidate exemplars are ranked using sentence embeddings based on trigger content, understanding statements, and question-answer text, and a small set of highly relevant examples is then sampled to provide few-shot guidance while preserving diversity. Conditioned on the retrieved exemplars and the structured evidence state, the language model Gemini-3.1-flash-lite-preview~\cite{gemini_api_2026} generates a response intent, a brief rationale, an ordered insight sequence containing one to three steps, a prohibited cognitive statement, and a safety note. The insight-generation prompt is provided in \autoref{fig:insight_generation}. The insight sequence allows semantic, somatic, and regulatory behaviors, and the prompting scheme further constrains the generated outputs to remain contextually appropriate, behaviorally plausible, and grounded in the available evidence.

\begin{figure}[!h]
    \centering
    \includegraphics[width=\linewidth]{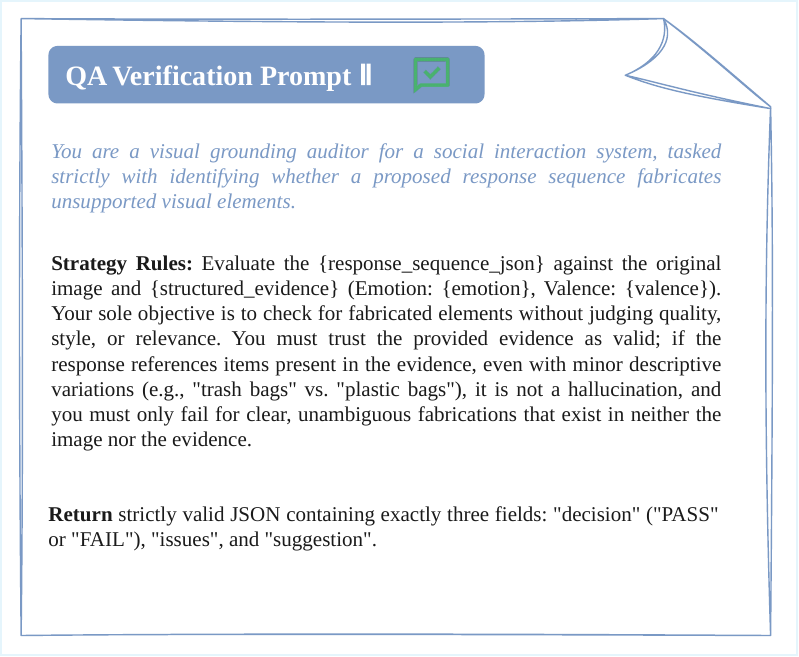}
    \caption{VLM Auditor Prompt.}
    \label{fig:qa_verification2}
\end{figure}

\noindent\textbf{Quality Control.}
To ensure annotation quality, the generated outputs are passed through a two-stage validation procedure. The first stage performs structural checking, verifying that the predicted intent belongs to the predefined set, that the insight sequence contains a valid number of steps, and that each step satisfies basic type and content constraints. Additional rules are introduced to remove formulaic semantic expressions, weak or non-informative somatic behaviors, and trivial rationales or safety notes. The second stage conducts visual auditing with Qwen3-VL-32B~\cite{bai2025qwen3} on the original image in order to detect fabricated people or major objects that are unsupported by both the image and the structured evidence. The visual auditing prompt is provided in \autoref{fig:qa_verification2}. To avoid unnecessary rejection, the auditor adopts a conservative decision policy and only marks a sample as failed when the unsupported content is clear. In the batch workflow, failed outputs may be regenerated and re-audited in subsequent cycles, while any remaining failures are reserved for final reporting and optional manual inspection.

\section{Annotation Examples}

\begin{figure*}[t]
  \centering
  \includegraphics[width=\linewidth]{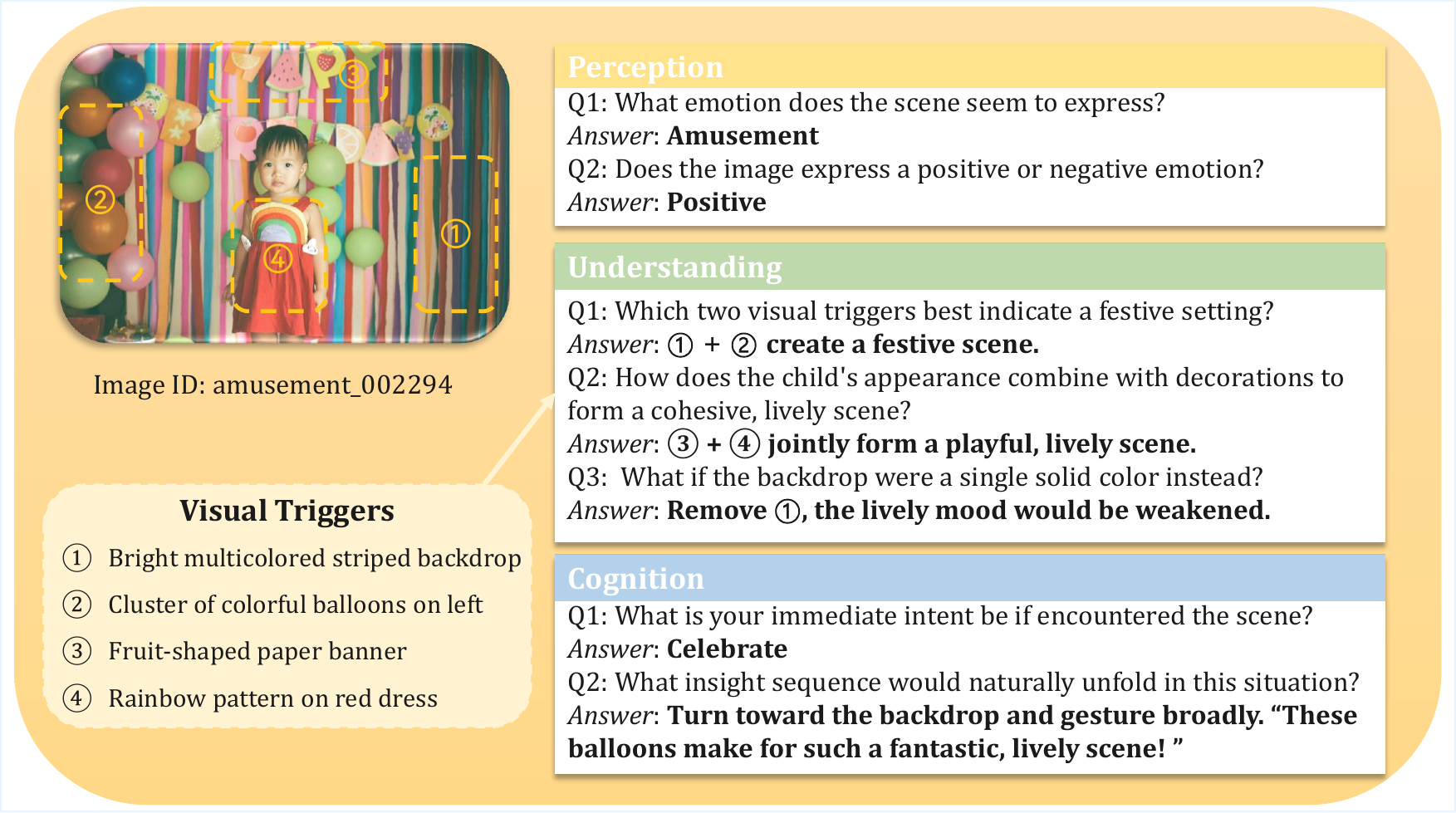}
  \caption{
  Hierarchical annotation example for amusement in InsightVQA.
  }
  \label{fig:amusement}
\end{figure*}

\begin{figure*}[h]
  \centering
  \includegraphics[width=\linewidth]{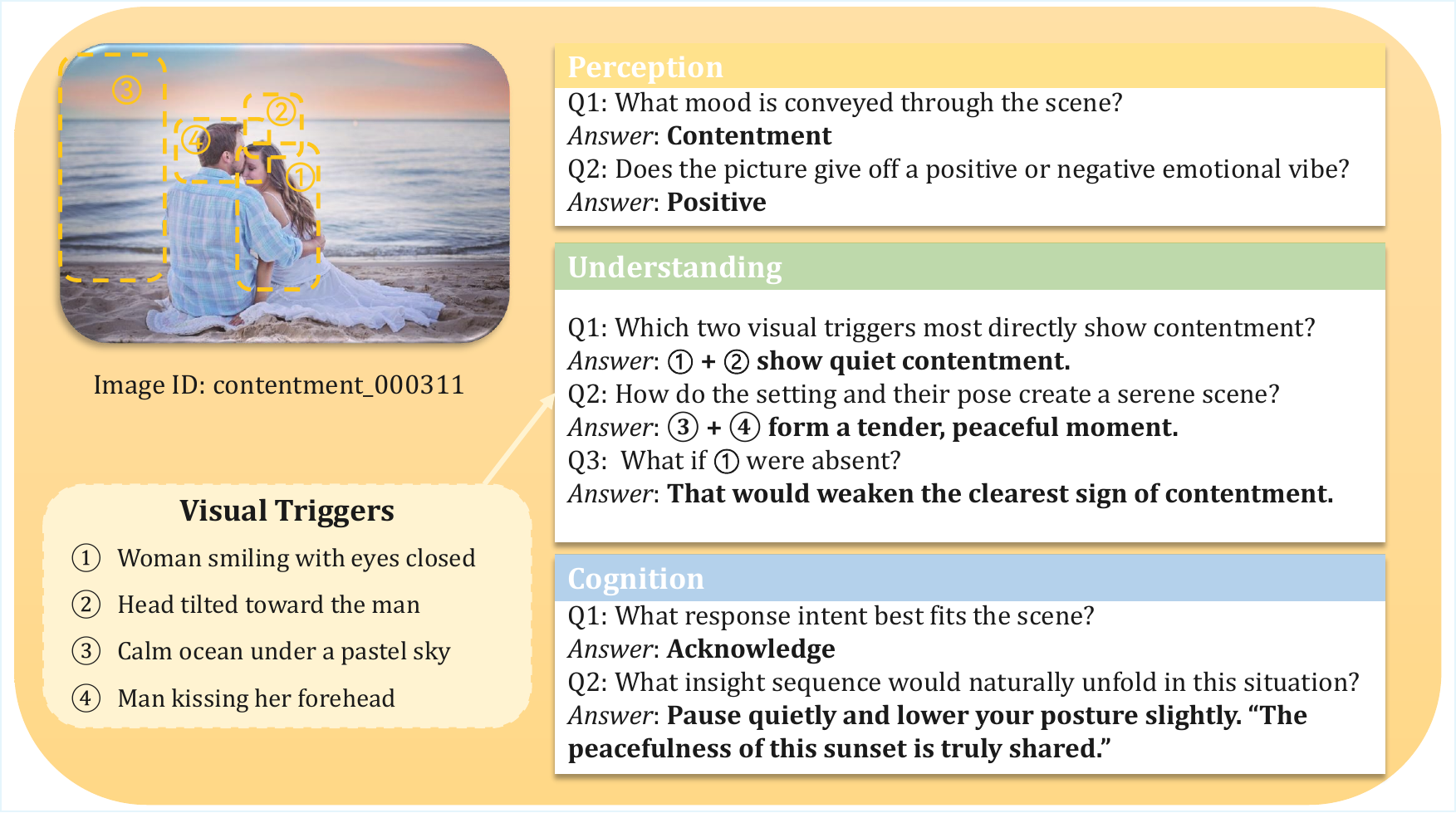}
  \caption{
  Hierarchical annotation example for \textbf{contentment} in InsightVQA.}
  \label{fig:contentment}
\end{figure*}

\begin{figure*}[!t]
  \centering
  \includegraphics[width=\linewidth]{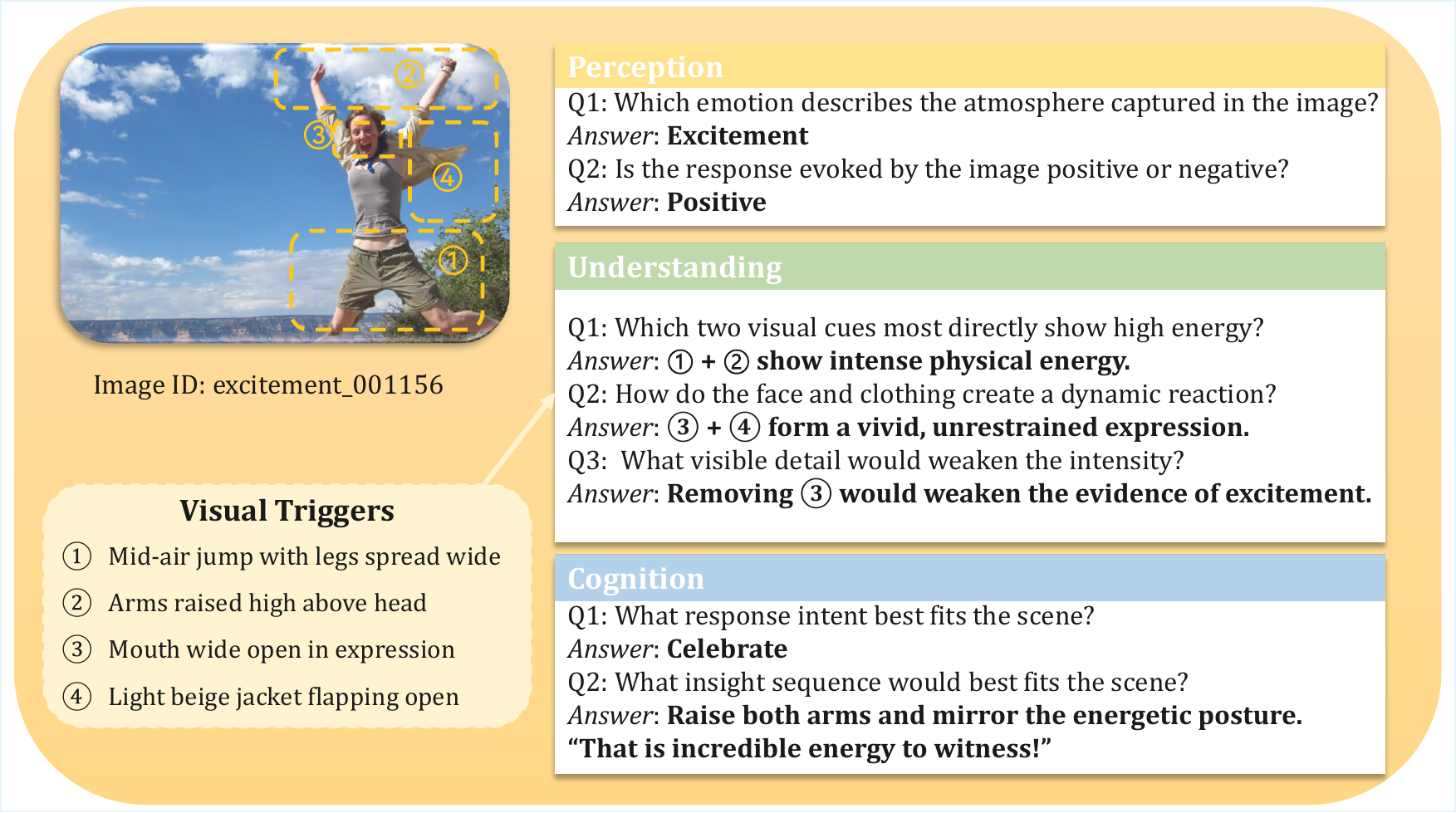}
  \caption{
  Hierarchical annotation example for \textbf{excitement} in InsightVQA.}
  \label{fig:excitement}
\end{figure*}

\begin{figure*}[!t]
  \centering
  \includegraphics[width=\linewidth]{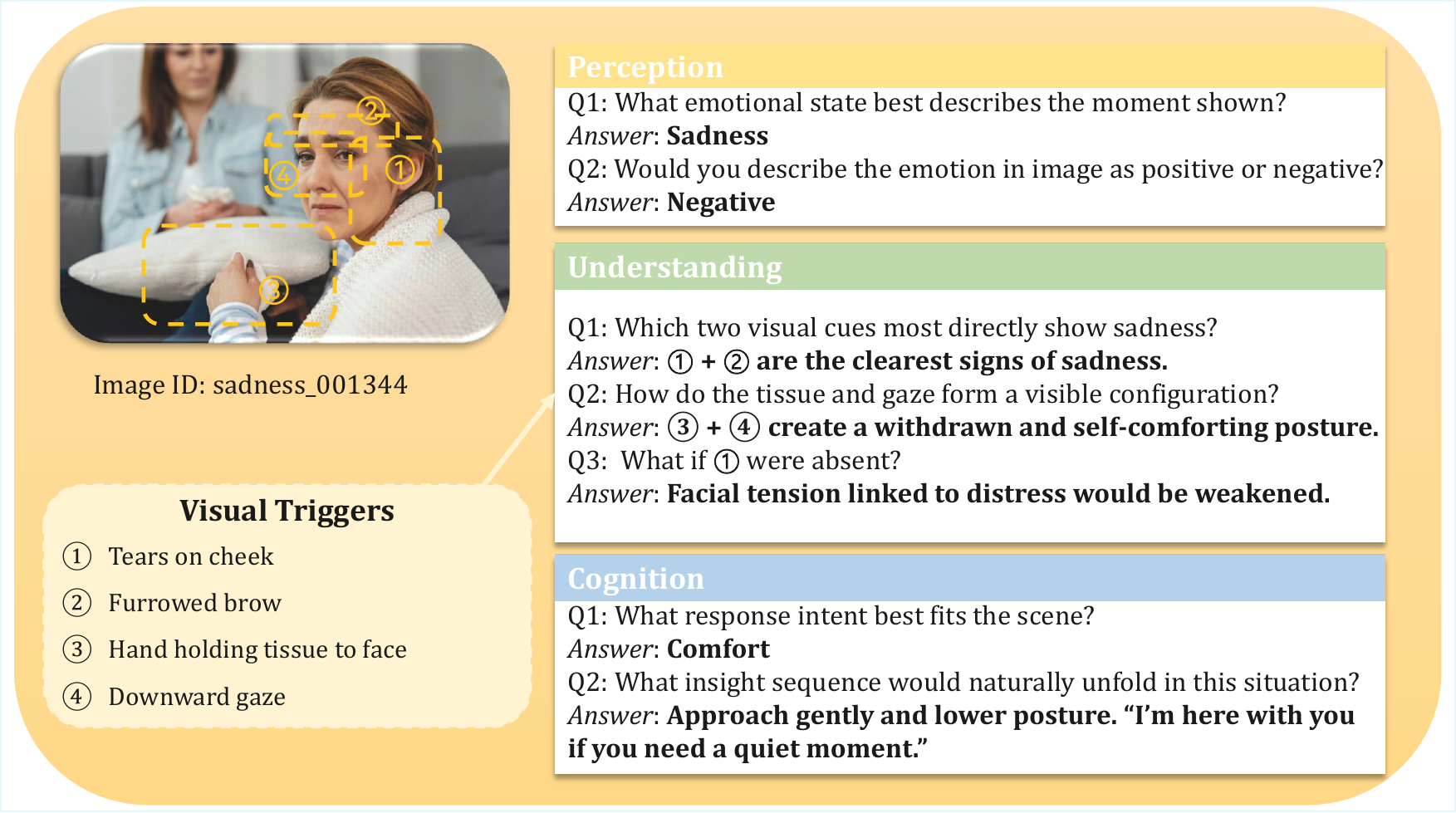}
  \caption{
  Hierarchical annotation example for \textbf{sadness} in InsightVQA.}
  \label{fig:sadness}
\end{figure*}

To illustrate how InsightVQA represents emotion understanding as a hierarchical process, we present four representative annotation examples covering amusement, contentment, excitement, and sadness, illustrated in Figures~\ref{fig:amusement}--\ref{fig:sadness}. Each example includes the input image with grounded visual triggers, followed by the corresponding annotations from the perception, understanding, and cognition layers. Together, these cases demonstrate how visible evidence is progressively transformed into emotion perception, evidence-grounded reasoning, and response-oriented cognition.

\section{Implementation Details}

\subsection{Detailed Information of Compared Models}

\noindent \textbf{QwenVL-2.5}~\cite{bai2025qwen2.5vl} is the latest flagship model in the Qwen vision-language series, featuring a redesigned native dynamic-resolution Vision Transformer (ViT) and Multimodal Rotary Position Embedding (MROPE) aligned with absolute time. This architecture enables the model to natively perceive spatial scales and temporal dynamics, achieving significant advancements in precise object localization, robust document parsing, and ultra-long video comprehension.

\noindent \textbf{Qwen3-Max}~\cite{bai2025qwen3} is the most powerful flagship vision-language model in the Qwen series to date, featuring a native multimodal pre-training approach that supports interleaved contexts of up to 256K tokens. Architecturally, it introduces key upgrades including an enhanced Interleaved-MROPE for superior spatial-temporal modeling, DeepStack integration for multi-level vision-language alignment, and explicit textual timestamp alignment for precise video grounding. By scaling from 2B to 235B parameters across dense and Mixture-of-Experts (MoE) variants, Qwen3-VL delivers state-of-the-art performance in complex reasoning, long-document parsing, and real-world agentic tasks, effectively bridging perception, reasoning, and action.

\noindent \textbf{LLaVA-OneVision-1.5}~\cite{xie025LLaVA-OneVision-1.5} is a novel family of large multimodal models that provides a fully open and efficient framework for democratized multimodal training. It optimizes vision-language alignment to significantly enhance capabilities in single-image understanding, multi-image reasoning, and video analysis.

\noindent \textbf{InternVL3.5}~\cite{wang2025internvl3} is a high-performance open-source multimodal model that significantly advances multidisciplinary reasoning and operational efficiency within the "ViT-MLP-LLM" paradigm. It introduces a Cascade Reinforcement Learning (Cascade RL) framework to bridge the gap between offline and online optimization, alongside architectural innovations like Vision Resolution Routing (ViR) and Decoupled Vision-Language Deployment (DVD) for enhanced inference speed.

\noindent \textbf{Deepseek-VL-chat}~\cite{lu2024deepseek} is an open-source Vision-Language model family specifically designed for practical, real-world applications. It features a hybrid vision encoder architecture that efficiently processes high-resolution inputs within a fixed token budget, ensuring high performance while maintaining low computational overhead. By integrating vision and language through a deep alignment pre-training strategy and fine-tuning on diverse, use-case-driven instruction data, the model excels in tasks like web screenshot analysis, OCR, and complex chart understanding.

\noindent \textbf{GPT-4o}~\cite{hurst2024gpt} is an end-to-end omni model capable of processing and generating any combination of text, audio, image, and video inputs and outputs. By utilizing a unified neural network architecture, it achieves real-time responsiveness with an average audio latency of 320 milliseconds, matching human conversational speeds. While maintaining parity with GPT-4 Turbo in English text and coding tasks, GPT-4o delivers significant advancements in non-English language support, visual understanding, and speech interaction, all while offering higher inference efficiency and lower API costs for complex multimodal applications.

\noindent \textbf{Gemini-2.5-flash}~\cite{comanici2025gemini} is a high-performance multimodal model designed to provide advanced reasoning and low-latency responses at a fraction of the compute requirements of larger models. As a key component of the Gemini 2.X family, it natively supports interleaved inputs of text, images, audio, and video while offering an expansive context window capable of processing up to 3 hours of video content.

\noindent \textbf{Claude-3.7-sonnet}~\cite{kasireddy2026evaluating} is Anthropic’s first hybrid reasoning model, designed to seamlessly integrate rapid responses with extended, step-by-step thinking within a single framework. By introducing a "Thinking Mode" that allows for serial test-time compute scaling, the model can adapt its depth of reasoning to solve complex multidisciplinary problems, particularly excelling in coding and advanced mathematical reasoning.

\noindent \textbf{EmoViT}~\cite{xie2024emovit} is a specialized multimodal large model designed for advanced emotion understanding. Built on the InstructBLIP architecture, it utilizes a parameter-efficient fine-tuning approach, updating only the instruction-aware Q-Former module while keeping both the visual encoder and large language model (LLM) frozen. This design enables precise alignment between visual signals and emotional semantics through diverse instruction-tuning data. As a result, EmoVIT excels at complex affective tasks, offering high instruction sensitivity and strong generalization across multiple benchmarks.

\noindent \textbf{Emotion-Qwen}~\cite{huang2025emotion} is a unified multimodal framework designed for both robust emotion understanding and general vision-language reasoning. Based on the Qwen2-VL architecture, it introduces a novel Hybrid Compressor leveraging a Mixture-of-Experts (MoE) structure to dynamically route inputs, effectively balancing emotion-specific processing with general multimodal capabilities to prevent catastrophic forgetting. Through a meticulously designed three-stage training pipeline—comprising vision-language alignment, emotion-specific knowledge enhancement, and all-task instruction tuning—the model achieves state-of-the-art performance in Video Emotion Recognition, affective VQA, and general visual understanding tasks.

\noindent \textbf{EmoCaliber}~\cite{wu2025emocaliber} is a multimodal framework designed to enhance the reliability of Visual Emotion Comprehension. By introducing a "confidence verbalization" mechanism, it enables the model to express its self-assessed certainty in natural language, addressing the inherent subjectivity of affective perception. Built on a three-stage training pipeline, EmoCaliber significantly improves model calibration and trustworthiness, providing dependable confidence signals alongside emotion predictions in complex visual scenarios.

\subsection{Detailed Evaluation Metrics}

To comprehensively evaluate model performance across different types of tasks, we employ a variety of metrics tailored to the nature of each task.

\noindent \textbf{Perception Evaluation.} Accuracy is used primarily for classification tasks. It measures the proportion of correctly predicted instances over the total number of instances. Formally, it is defined as:

$$ \text{Accuracy} = \frac{N_\text{correct}}{N_\text{total}} $$

\noindent where $N_\text{correct}$ denotes the number of correctly predicted samples and $N_\text{total}$ is the total number of samples.

\noindent \textbf{Understanding Evaluation.} For open-ended or generation tasks, simple accuracy may not fully capture performance, so we use precision, recall, and F1-score:

\begin{itemize} [leftmargin=*]
    \item \textbf{Precision} measures the proportion of correctly predicted positive instances among all predicted positives:
    $$
    \text{Precision} = \frac{\text{TP}}{\text{TP + FP}}
    $$
    \item \textbf{Recall} measures the proportion of correctly predicted positive instances among all actual positives:
    $$
    \text{Recall} = \frac{\text{TP}}{\text{TP + FN}}
    $$
    \item \textbf{F1-score} is the harmonic mean of precision and recall, providing a balanced measure:
    $$
    \text{F1} = 2 \cdot \frac{\text{Precision} \cdot \text{Recall}}{\text{Precision} + \text{Recall}}
    $$
\end{itemize}

\noindent where TP, FP, and FN denote true positives, false positives, and false negatives, respectively. These metrics are especially useful when dealing with imbalanced datasets or tasks with multiple valid outputs.

\noindent \textbf{Cognition Evaluation.} For tasks that involve ranking or ordering items, we employ ranking-based evaluation metrics:

\begin{itemize} [leftmargin=*]
    \item \textbf{Top-1 Accuracy} measures the fraction of instances where the top-ranked prediction matches the ground truth:
    $$
    \text{Top-1 Accuracy} = \frac{N_\text{top1 correct}}{N_\text{total}}
    $$
    \item \textbf{Spearman Rank Correlation} evaluates how well the predicted ranking of items correlates with the true ranking. It is defined as:
    $$
    \rho = 1 - \frac{6 \sum_{i=1}^{n} d_i^2}{n(n^2-1)}
    $$
    where $d_i = R_i^\text{gold} - R_i^\text{pred}$ is the difference between the predicted and true ranks of item $i$, and $n$ is the number of items. A higher Spearman correlation indicates better agreement between the predicted and true rankings.
\end{itemize}

By employing this set of complementary metrics, we ensure that model evaluation captures multiple aspects of performance, from classification accuracy to ranking fidelity and output quality in open-ended scenarios.

\end{document}